\documentclass[runningheads]{llncs}

\makeatletter
\newcommand{\@chapapp}{\relax}
\makeatother
\usepackage{graphicx}
\usepackage[frozencache=true,cachedir=.]{minted}
\usepackage[T1]{fontenc}
\usepackage{dirtytalk}
\usepackage{amsmath}
\usepackage{gensymb}
\usepackage{enumitem}
\usepackage{float}
\usepackage{caption}
\usepackage{subcaption}
\usepackage[ruled,vlined]{algorithm2e}
\usepackage{multicol}
\usepackage{booktabs}
\usepackage{multirow}
\usepackage{siunitx}

\usepackage[final]{hyperref}
\hypersetup{colorlinks,allcolors=black}
\usepackage{tikz}
\def\checkmark{\tikz\fill[scale=0.4](0,.35) -- (.25,0) -- (1,.7) -- (.25,.15) -- cycle;}

\usepackage{sectsty}
\subsectionfont{\fontsize{12}{15}\selectfont}

\usepackage[style=nature,backend=bibtex,urldate=comp]{biblatex}
\usepackage{appendix}
\usepackage[capitalise,nameinlink,noabbrev]{cleveref}
\crefname{appendixfigure}{Supplementary Figure}{Supplementary Figures}
\crefname{appendixtable}{Supplementary Table}{Supplementary Tables}
\crefname{appendixname}{Supplementary Information}{Supplementary Information}

\newcolumntype{d}[1]{D{.}{.}{#1}}
\SetArgSty{textnormal}

\usepackage{geometry}
\begin{document}
%TC:ignore
\title{Scaling up DNA digital data storage by efficiently predicting DNA hybridisation using deep learning}
\author{\textbf{David Buterez}\thanks{Work conceptualised and partially implemented while at Imperial College London.}}
\authorrunning{D. Buterez}
\institute{Department of Computer Science and Technology \\
  University of Cambridge, Cambridge, UK\\}

\titlerunning{Published in \textit{Nature Scientific Reports} (\href{https://doi.org/10.1038/s41598-021-97238-y}{https://doi.org/10.1038/s41598-021-97238-y})}
\maketitle              % typeset the header of the contribution
%
%TC:endignore
%TC:ignore
% \linenumbers
\begin{abstract}
  Deoxyribonucleic acid (DNA) has shown great promise in enabling computational applications, most notably in the fields of DNA digital data storage and DNA computing. Information is encoded as DNA strands, which will naturally bind in solution, thus enabling search and pattern-matching capabilities. Being able to control and predict the process of DNA hybridisation is crucial for the ambitious future of \textit{Hybrid Molecular-Electronic Computing}. Current tools are, however, limited in terms of throughput and applicability to large-scale problems.
  \\\\
  We present the first comprehensive study of machine learning methods applied to the task of predicting DNA hybridisation. For this purpose, we introduce an \textit{in silico}-generated hybridisation dataset of over 2.5 million data points, enabling the use of deep learning. Depending on hardware, we achieve a reduction in inference time ranging from one to over two orders of magnitude compared to the state-of-the-art, while retaining high fidelity. We then discuss the integration of our methods in modern, scalable workflows.

\end{abstract}
%TC:endignore
The use of DNA to facilitate computation is an active area of research, dating back to 1994 when Leonard Adleman solved a seven-node instance of the Hamiltonian path problem, using the toolbox of (DNA) molecular biology \cite{adleman}. This pioneering work opened the gate to many interesting questions: can molecular machines be used to solve intractable problems? Is DNA suitable for long-term storage of digital information? More recently, are such methods scalable in the era of Big Data?
\\\\
The focus has gradually changed from solving difficult computational problems to exploiting desirable properties of DNA, leading to the development of \emph{DNA digital data storage}. It is now generally accepted that the amount of digital data is doubling at least every two years. Predictions from Seagate estimate that this quantity, called the \textit{Global Datasphere}, will grow from 33 ZB (zettabytes) in 2018 to 175 ZB by 2025 \cite{Reinsel2018TheCore}. Furthermore, the dominant storage mediums are traditional, with 59\% of the storage capacity expected to come from hard disk drives and 26\% from flash technologies. Synthetic DNA has been argued to be an attractive storage medium, for at least three prominent reasons \cite{Carmean}:

\begin{enumerate}[itemsep=4pt]
  \item \textbf{Density} -- The theoretical maximum information density of DNA is $10^{18}$ $\text{B/mm}^3$. Comparatively, this is a 7 orders of magnitude increase over tape storage.

  \item \textbf{Durability} -- Depending on storage conditions, DNA can be preserved for at least a few hundred years. Fossil studies reveal that DNA has a half-life of 521 years \cite{Allentoft2012TheFossils}. In appropriate conditions, researchers estimate that digital information can be recovered from DNA stored at the Global Seed Vault (at -18\degree C) after over 1 million years \cite{Grass2015RobustCodes}.

  \item \textbf{Future-proofing} -- Next-generation sequencing and technologies such as Oxford Nanopore \cite{Jain2016} have made reading and writing DNA more accessible than ever. Furthermore, since DNA is the fundamental building block of life, it will be relevant for as long as humans exist.
\end{enumerate}

\noindent In silicon-based computers, information is loaded and stored from unique locations denoted by a numerical address or index (random access). Implementing this property in DNA storage applications requires a short, arbitrary sequence of DNA acting as the index, enabling selection and amplification by polymerase chain reaction (PCR) \cite{Appuswamy2019OligoArchive:Hierarchy}, \cite{Goldman2013}. Using this design, the index DNA sequence acts as both a unique identifier and a PCR primer, thus it must fulfil standard primer length requirements, often between 18 and 26 nucleotides \cite{Rychlik2000}, and annealing and melting temperature requirements, often between 30\textdegree C and 65\textdegree C \cite{Yang2006}, \cite{pmid12045152}, \cite{BUSTIN201719}. As the scale of the primer library increases, designing sequences that strongly interact with their desired targets and weakly (preferably never) interact with non-intended sequences becomes progressively more difficult. Such sequence libraries are called \textbf{orthogonal}. It is also possible to exploit non-precise hybridisation to enable fuzzy or similarity search, by encoding closely-related entities (by some distance metric) in DNA, leading to \textbf{similar} datasets.
\\\\
Successful and precise DNA hybridisation is required to implement computational features directly at the level of DNA. The OligoArchive project \cite{Appuswamy2019OligoArchive:Hierarchy} introduced a technique capable of performing SQL operations such as selection, projection and join directly on the DNA molecules, by means of hybridisation and finding appropriate encodings for primary keys and attributes. Stewart et al. \cite{Stewart2018AEncodings} used an approximation for thermodynamic analysis that is differentiable (necessary for backpropagation) in the form of a modified sigmoid function, with the goal of creating a content-addressable DNA database. Unfortunately, the estimation is far from perfect and the authors remarked that an important future direction is a more accurate approximation for thermodynamic yield.
\\\\
Zhang et al. \cite{Zhang2018Hyb} predicted hybridisation kinetics from 36-nucleotide sequence pairs derived from human genes (non-synthetic), on a much more limited scale (100 pairs) using non-neural methods. We first introduced the concept of using CNNs to predict DNA hybridisation as a poster in June 2019, at the \textit{25th International Conference on DNA Computing and Molecular Programming} (DNA25), reporting significant gains in inference time. Close to our research direction, in a recent preprint Bee et al. \cite{Bee2020.05.25.115477} present a similarity search DNA database that employs a hybridisation predictor, a CNN trained on one-hot encodings of DNA. The authors state that hybridisation yields are either close to 0 or close to 1, suggesting that they do not consider intermediate values during training.
\\\\
Unfortunately, NUPACK, the state-of-the-art in the analysis and design of nucleic acid systems \cite{https://doi.org/10.1002/jcc.21596}, cannot be used for large-scale applications. Firstly, while NUPACK employs efficient dynamic programming algorithms, these still require $O(N^4)$ time and $O(N^2)$ memory, rendering multi-million applications intractable. Secondly, NUPACK 3 is only available as a collection of standalone executables, massively slowing down the performance due to repeated I/O operations. Thirdly, and perhaps most importantly, the computation of NUPACK is not differentiable, thus it cannot be used in neural networks. We acknowledge that the very recently introduced NUPACK 4 \cite{doi:10.1021/acssynbio.9b00523} has completely removed the second drawback. However, while NUPACK 4 provides faster, vectorised operations, the first point is only partially mitigated, as the massive speed improvements only apply to large complexes of long sequences. The third point still applies in full.
\\\\
We recognise the need for scalable and accurate prediction of DNA hybridisation, necessarily as a differentiable operation that can be seamlessly plugged into neural network models. Hybridisation of two single-stranded sequences is quantified by the equilibrium concentration of the DNA duplex, here referred to as \emph{yield}. We propose a modern, scalable deep learning approach for predicting DNA hybridisation (schematically illustrated in \Cref{fig:main-figure}), designed and evaluated on an \textit{in silico}-generated hybridisation dataset of over 2.5 million single-stranded DNA sequence pairs with ground truth yields computed at $37.0\degree, 42.0\degree, 47.0\degree, 52.0\degree, 57.0\degree,$ and $62.0\degree$ (Celsius), capturing a wide range and largely mirroring the temperatures used in \cite{Beliveau2018OligoMinerProbes}, \cite{Zhang2018Hyb}, making the following contributions:

\begin{figure}[H]
  \centering
  \captionsetup{justification=justified, singlelinecheck=off}
  \includegraphics[width=0.8\textwidth]{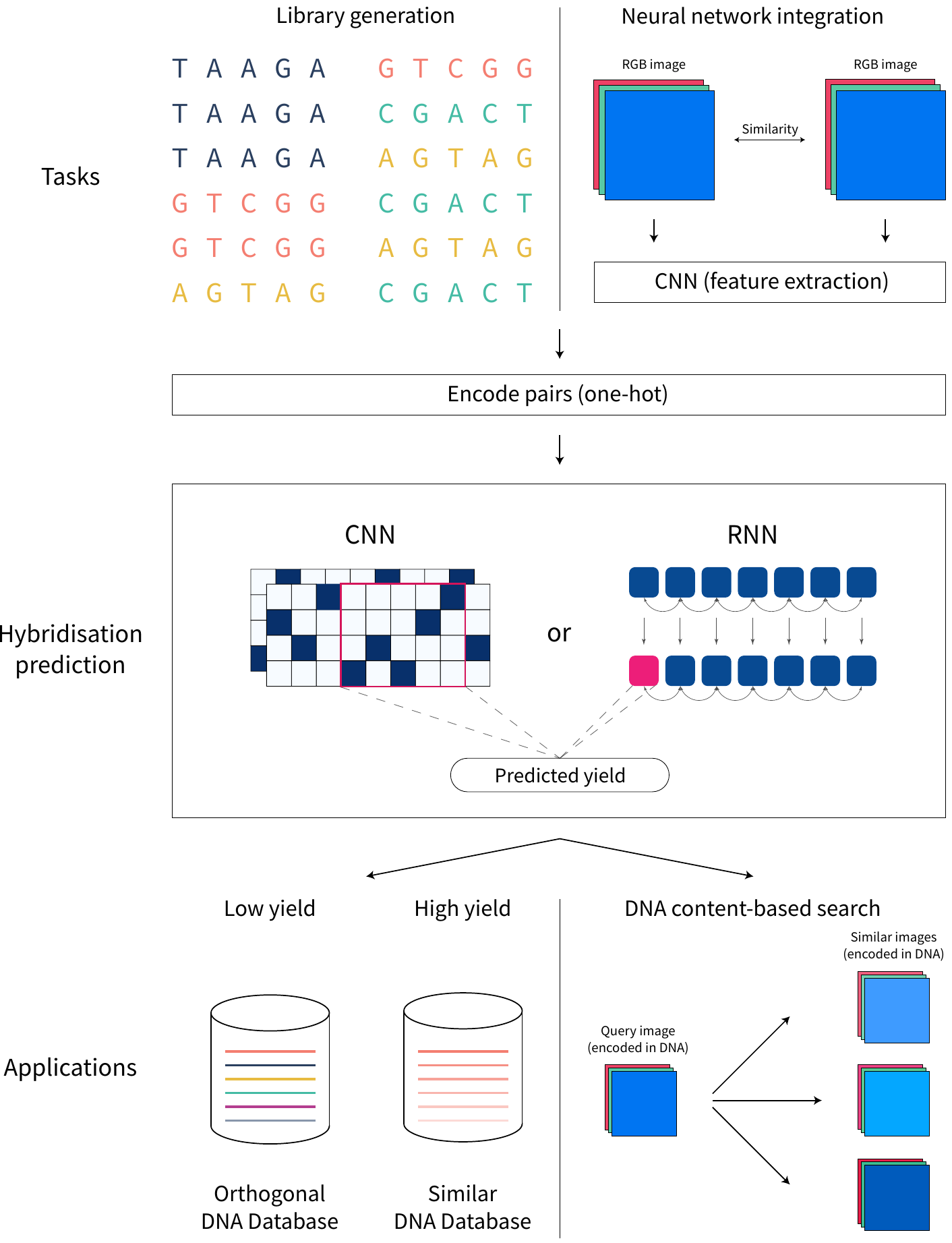}
  \caption{A high-level overview on how to integrate hybridisation prediction into DNA storage workflows. A trained machine learning model can be used as a standalone tool to assemble orthogonal or similar libraries of DNA sequences (left half of the figure). Alternatively, the neural network can be seamlessly integrated into a larger machine learning model as a subcomponent. The presented example is content-based search in a DNA database, where document features are extracted by a neural network (CNN for images, but text, video or audio inputs are conceivable) in a pairwise manner, another neural component generates appropriate encodings (usually one-hot) and the hybridisation predictor outputs the expected yield of the pair. Such a model is trained to associate similar documents to similar single stranded DNA sequences that form stable duplexes with the query sequence (right half of the figure).}
  \label{fig:main-figure}
\end{figure}

\begin{enumerate}[itemsep=4pt]
  \item \textbf{Integration with neural networks} -- We investigate Convolutional Neural Networks (CNNs), Recurrent Neural Networks (RNN) and Transformers. Naturally, we formulate the problem such that backpropagation can be applied.
  \item \textbf{Speed and accuracy} -- A decrease in inference time of one to over two orders of magnitude compared to non-neural methods, while we empirically show that our best models can reach over 0.965 in $\text{F}_1$ score per class after binarising the predicted yields.
\end{enumerate}

\section*{Results}
\subsection*{Designing a diverse hybridisation dataset}
\label{subsec:design_dataset}
\begin{figure}[H]
  \centering
  \captionsetup{justification=centering}
  \begin{subfigure}[t]{0.5\textwidth}
    \includegraphics[width=\textwidth]{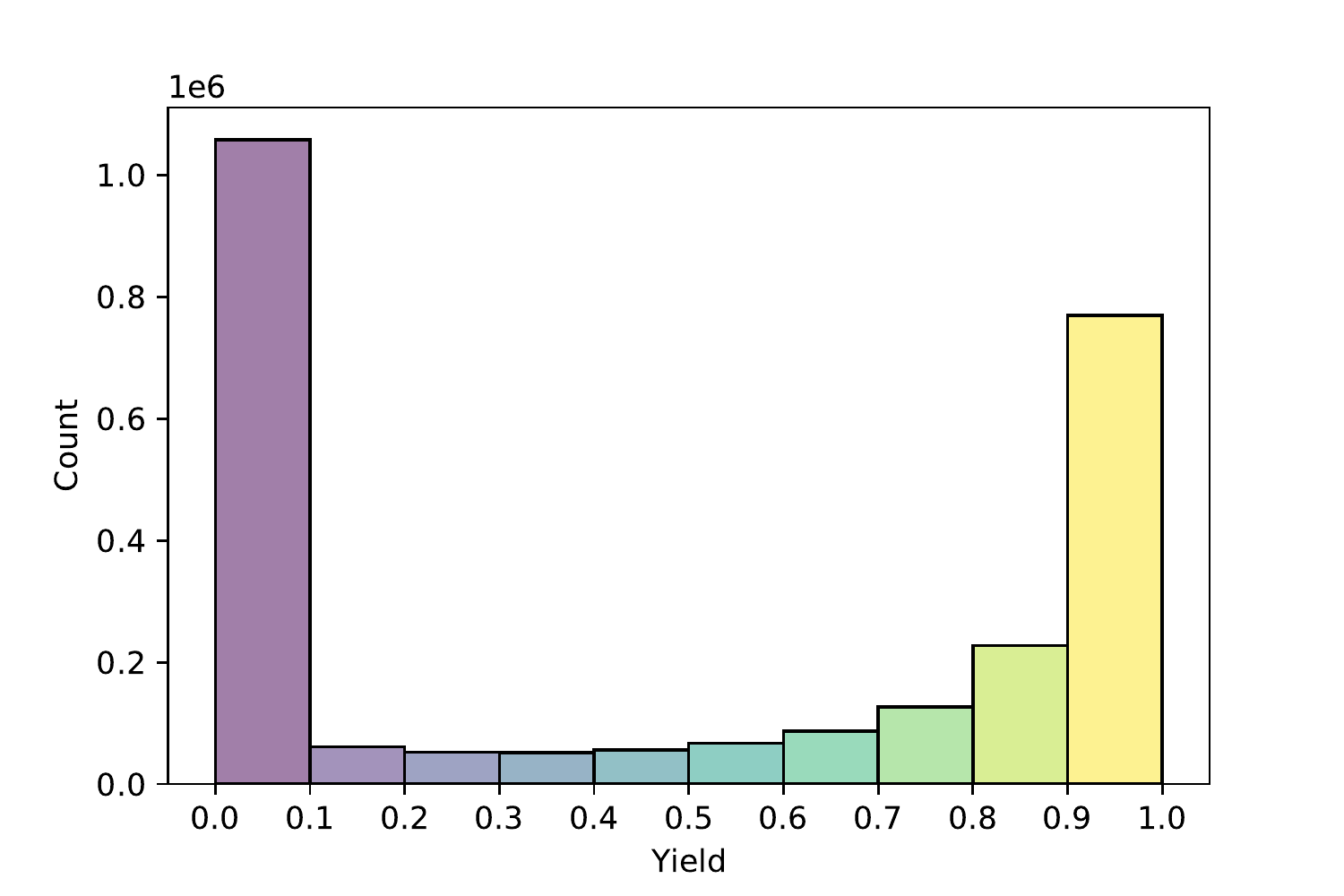}
    \caption{Histogram of the hybridisation dataset.}
    \label{fig:yield_hist}
  \end{subfigure}\hfill
  \begin{subfigure}[t]{0.5\textwidth}
    \includegraphics[width=\textwidth]{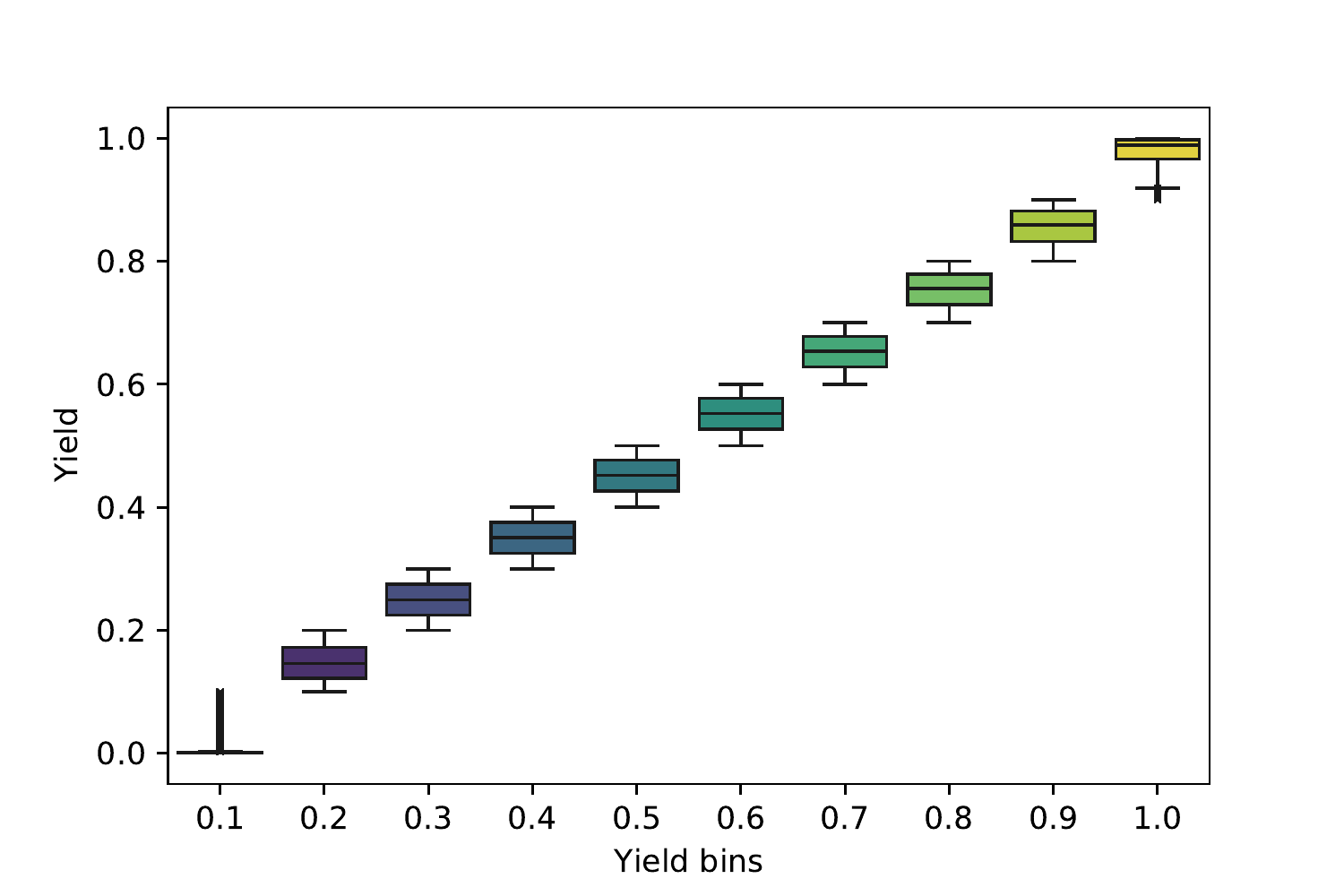}
    \caption{Box plot of the 10 yield bins (\textit{x axis} labelled by the right endpoint; means not shown).}
    \label{fig:yield_box_plot}
  \end{subfigure}
  \captionsetup{justification=justified, singlelinecheck=off}
  \caption{Visual summary of the yield distribution at 57\textdegree C (the temperature used throughout the paper). For a discussion on the behaviour of the yield at different temperatures, see \Cref{apx:sec-yield-temps}. The yield is binned in 10 groups, each spanning a 0.1 interval. Brighter colours correspond to higher yield and are shared for the two subfigures. \textbf{a}, Low and high values are the most numerous, which is expected considering our generative procedure. \textbf{b}, The highest density is achieved at the extremes of 0, respectively 1. Given how sensitive the molecules are even to a 1-base change, we count the entire intermediate range of yields $[0.1, 0.9)$ as one entity when considering how balanced the dataset is. In this regard, \num[group-separator={,}]{1058364} pairs achieve low yields ($< 0.1$), \num[group-separator={,}]{769750} achieve very high yields ($\geq 0.9$) and \num[group-separator={,}]{728862} are in-between. It is important that samples with extremely low or high yields are well represented. In particular, there are virtually endless combinations of base pairs resulting in minimum yield.}
  \label{fig:yield_distribution}
\end{figure}

To achieve a high-quality, comprehensive hybridisation dataset, the following two criteria must be satisfied: \textbf{size} -- large enough to allow a broad range of machine learning methods (including deep learning) and \textbf{variety} -- capture as many DNA interactions as possible, not only those corresponding to maximum or minimum yield. Unfortunately, we are not aware of any hybridisation dataset of biological origin that satisfies these conditions. Therefore, we introduce a synthetic dataset, where the DNA sequence pairs are generated from scratch and the yield computations are carried out by NUPACK.
\\\\
Following the procedures presented in \textit{Methods}, a final dataset of \num[group-separator={,}]{2556976} sequence pairs or just over 2.5 million data points is assembled, with sequence lengths varying in the range 18--26. By examining the properties of extreme and intermediary samples, a machine learning model should ideally be able to deduce the Watson-Crick base-pairing rules and how these affect the yield. The yield distribution is summarised visually in \cref{fig:yield_distribution}. For evaluation, we used stratified splits of \num[group-separator={,}]{2045579}  for training, \num[group-separator={,}]{255696} for validation and \num[group-separator={,}]{255701} for testing.

\captionsetup[subfigure]{labelformat=simple}

\noindent To establish a baseline and study the difficulty of the task, we extend the dataset with extracted features for each sequence, enabling classical machine learning techniques to be applied. Alignment scores reveal the relatedness of two biological sequences and are hypothesised to be good predictors of DNA annealing even for arbitrary, artificially-generated sequences that are much shorter than typical naturally-occurring DNA fragments. As such, the first extracted feature is given by alignment scores, computed according to \textit{Methods}. Furthermore, calculations such as the minimum free energy (MFE) of secondary structures indicate potentially undesirable behaviour. Motivating examples are provided in \cref{fig:npk_all}.
\\\\
Overall, we extend the dataset with the alignment score, secondary structure MFE, concentration of the single-stranded and double-stranded forms, and the GC content percentage of each sequence (ablation study in \cref{apx:abl-study}).
\subsection*{Supervised learning using extracted features}
\label{subsec:simple_ml}
The choice of machine learning algorithms is motivated by previous work (LDA in \cite{Beliveau2018OligoMinerProbes}), scalability to large problems (millions of samples) and widespread adoption and recognition. The four algorithms we evaluate are Linear Discriminant Analysis (LDA), Quadratic Discriminant Analysis (QDA), Random Forests (RF) and Neural Networks (NN). Our criteria prohibit the use of some well-known algorithms such as support-vector machines (SVM) due to limited scalability. For all chosen methods we perform hyperparameter optimisation, with the exception of QDA (very small number of tunable hyperparameters). For additional details see \cref{apx-sec:simple_ml_hyperparam}.
\\\\
The results of \cref{fig:simple_ml_results}, \ref{tab:deep_learning_metrics} confirm that random forests and neural networks are the top performers on this classification task. Furthermore, the more granular performance report from \cref{fig:simple_ml_results} indicates that all chosen methods perform well on the classification task. A general trend, consistent across all algorithms, can be identified. Precision for the \textbf{Low} class is high, indicating that when a low prediction is made, it is very likely correct. However, the recall for this class is relatively low, signalling that \textbf{Low} samples are often classified incorrectly as \textbf{High}. Almost all members of \textbf{High} are identified correctly (high recall). This trend is at its extreme for LDA, with the others (QDA, RF, NN) becoming more and more balanced across the two classes.
\\\\
We acknowledge two important limitations of this approach. Firstly, \textbf{generality} -- ideally, yields should be approximated directly, as a regression task, since different applications might require different cut-offs for the classification labels. Unfortunately, the methods presented so far are unable to approximate yields directly on our challenging dataset. Secondly, \textbf{performance and scalability} -- while the actual inference times of the algorithms in \cref{fig:simple_ml_results} are low, predictions require 9 pre-computations. Preferably, ML methods should be applied directly to sequences.
\\\\
To address these limitations we introduce the use of deep learning methods to predict DNA hybridisation directly from pairs of sequences.

\subsection*{Deep learning strategies for DNA hybridisation}
\label{subsec:deep_learning}
We employ two different deep learning paradigms: image-based, or more generally, models that operate on 2-dimensional grids with explicitly defined coordinates, and sequence-based. For the first category, we choose Convolutional Neural Networks (CNNs) as a representative, while for sequence models we investigate both Recurrent Neural Networks (RNNs) and Transformers.
\\\\
One of the first challenges in designing CNN models is to find a 2-dimensional representation of DNA sequences that enables convolutions. Our inputs are pairs of single-stranded DNA represented using the alphabet $\{\text{A}, \text{C}, \text{G}, \text{T}\}$. The proposed 2D representation is as $4 \times N \times 2$ images ($\textit{height} \times \textit{width} \times \textit{channels}$), where $N$ is the maximum sequence length.

\begin{figure}[H]
  \centering
  \captionsetup{justification=justified, singlelinecheck=off}
  \begin{subfigure}[t]{0.3\textwidth}
    \includegraphics[width=\textwidth]{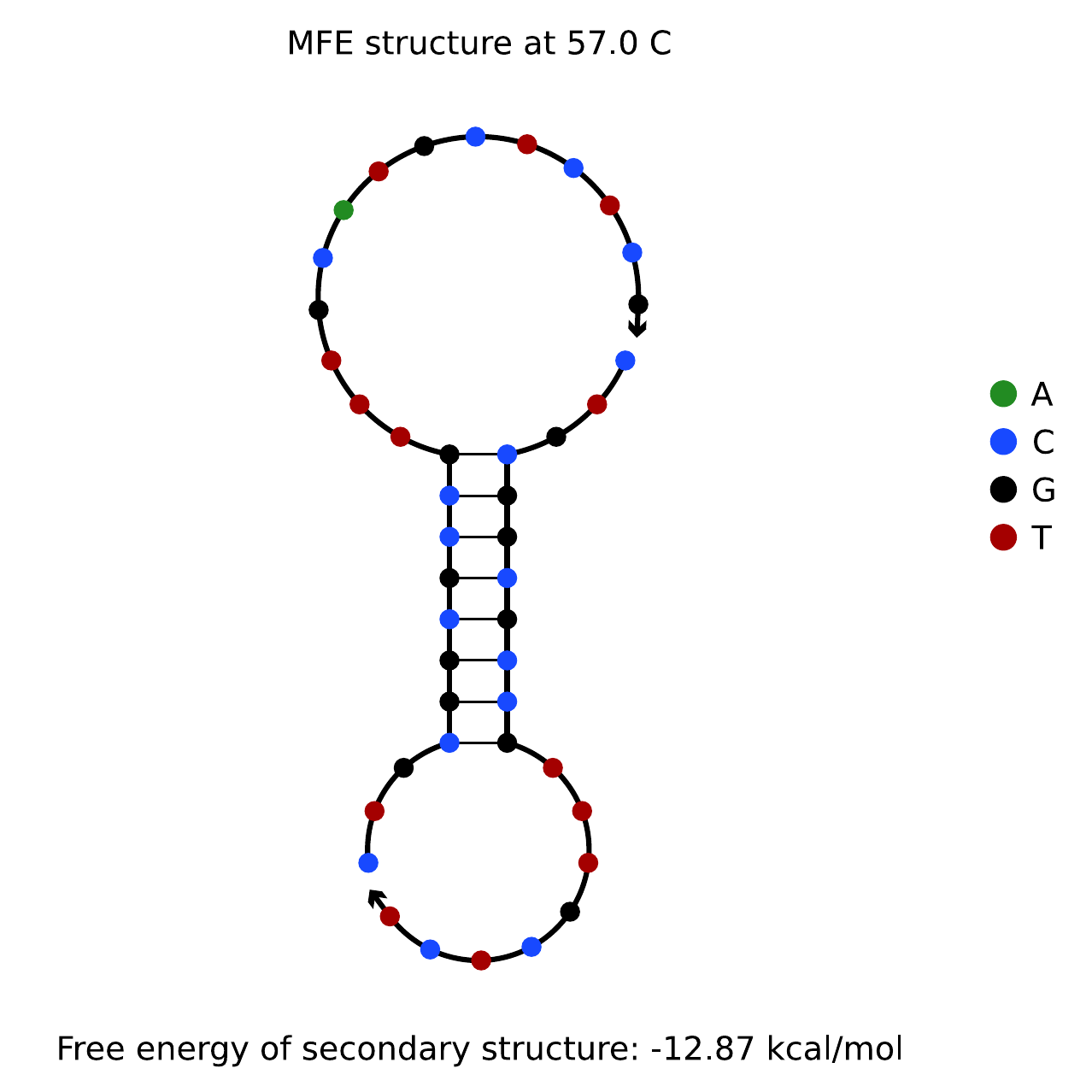}
    \caption{Pair with alignment score 31.}
    \label{fig:npkaln2}
  \end{subfigure}
  \hspace{60pt}
  \begin{subfigure}[t]{0.3\textwidth}
    \includegraphics[width=\textwidth]{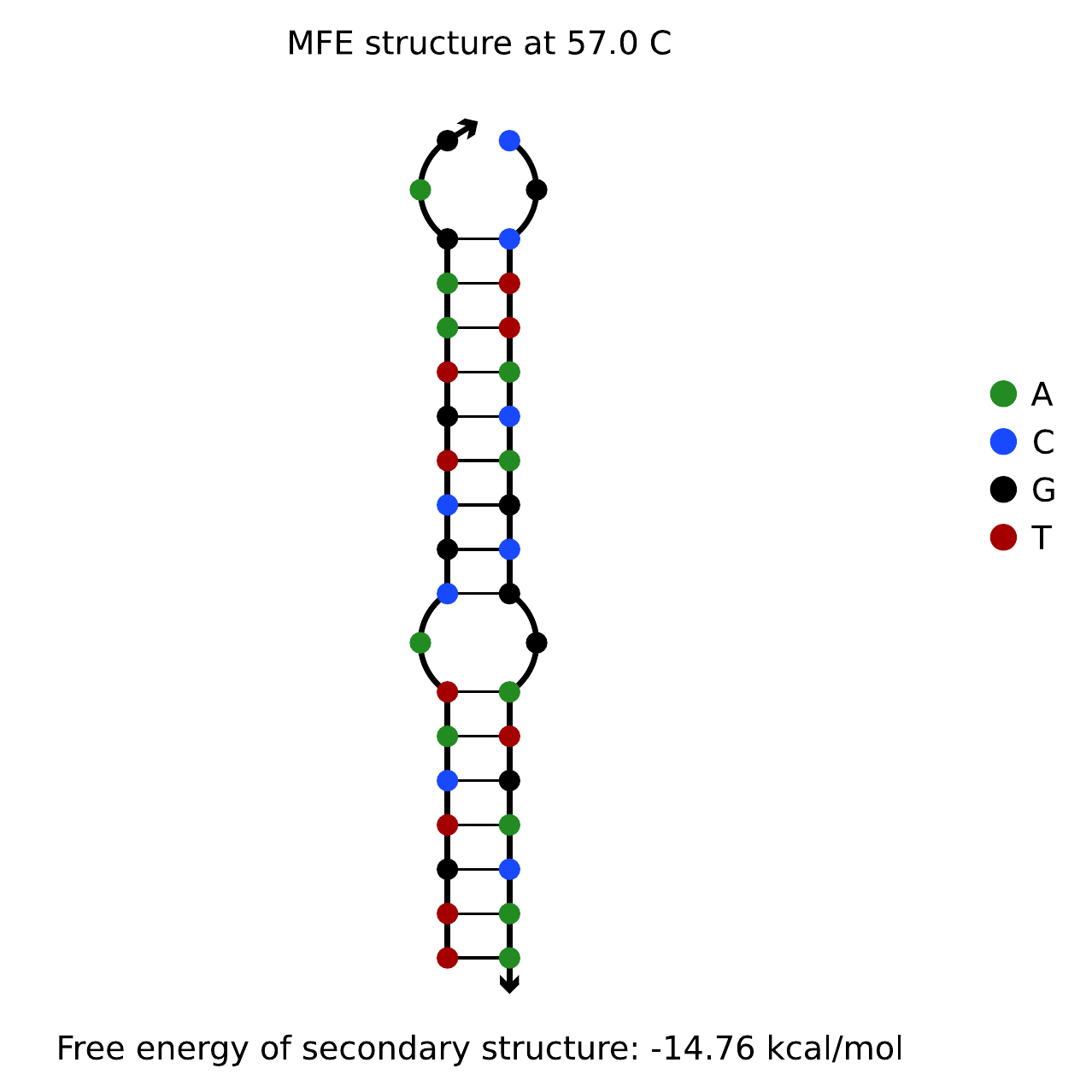}
    \caption{Pair with alignment score 77.}
    \label{fig:npkaln1}
  \end{subfigure}
  \vspace{20pt}
  \begin{subfigure}[t]{0.3\textwidth}
    \includegraphics[width=\textwidth]{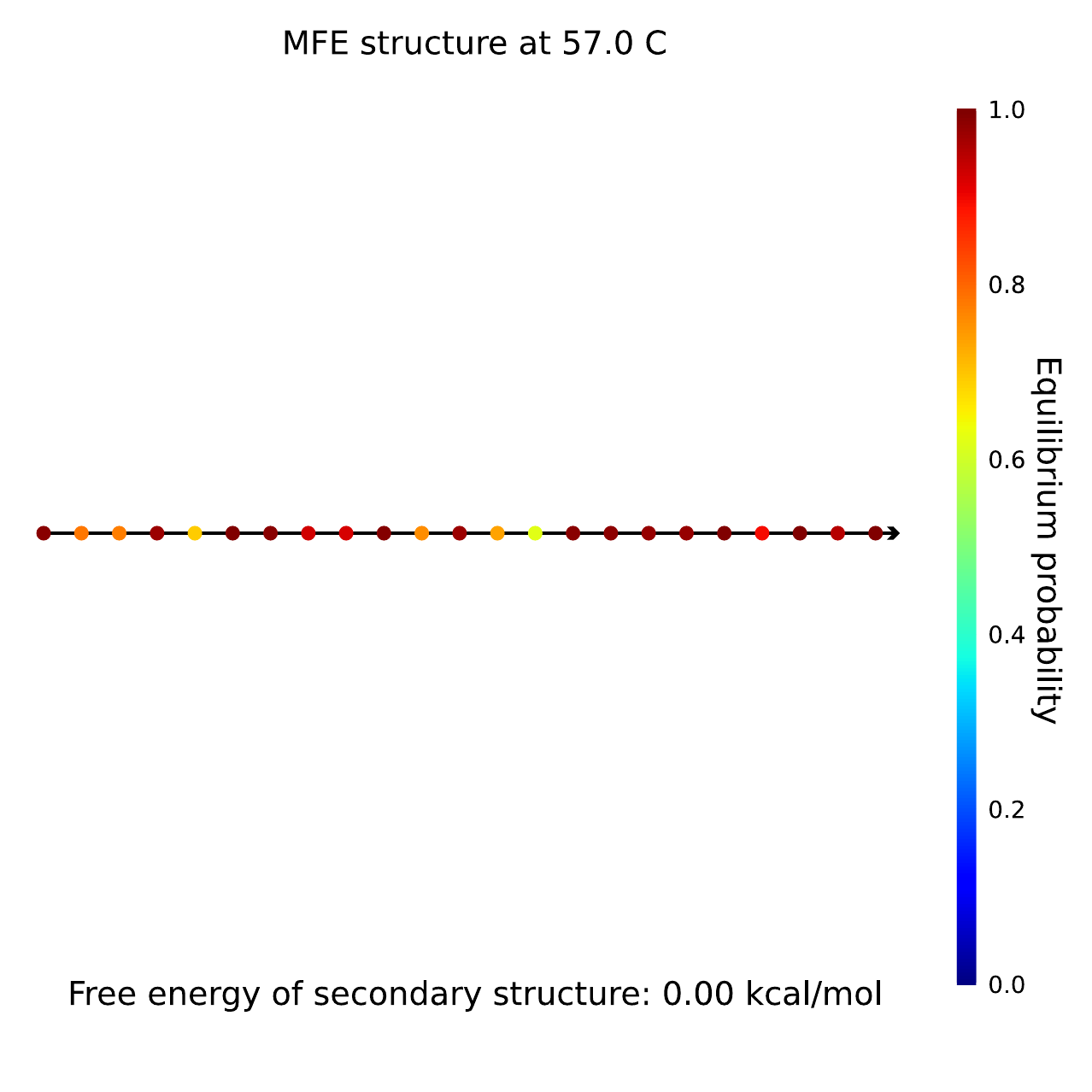}
    \caption{A sequence with no remarkable secondary structure.}
    \label{fig:singlemfe1}
  \end{subfigure}
  \hspace{60pt}
  \begin{subfigure}[t]{0.3\textwidth}
    \includegraphics[width=\textwidth]{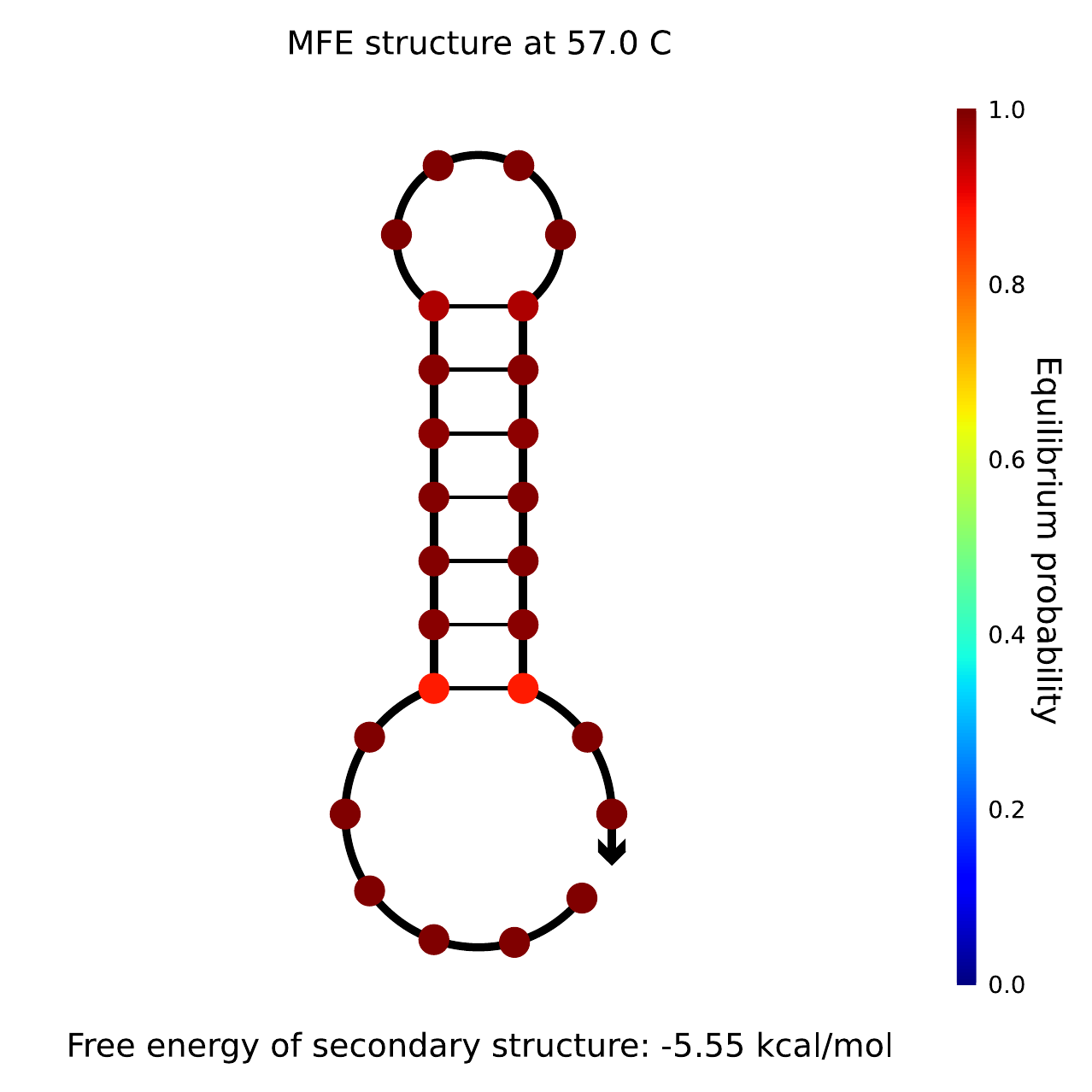}
    \caption{A sequence that exhibits self-complementarity.}
    \label{fig:singlemfe2}
  \end{subfigure}
  \vspace{20pt}

  \begin{subfigure}[t]{0.3\textwidth}
    \includegraphics[width=\textwidth]{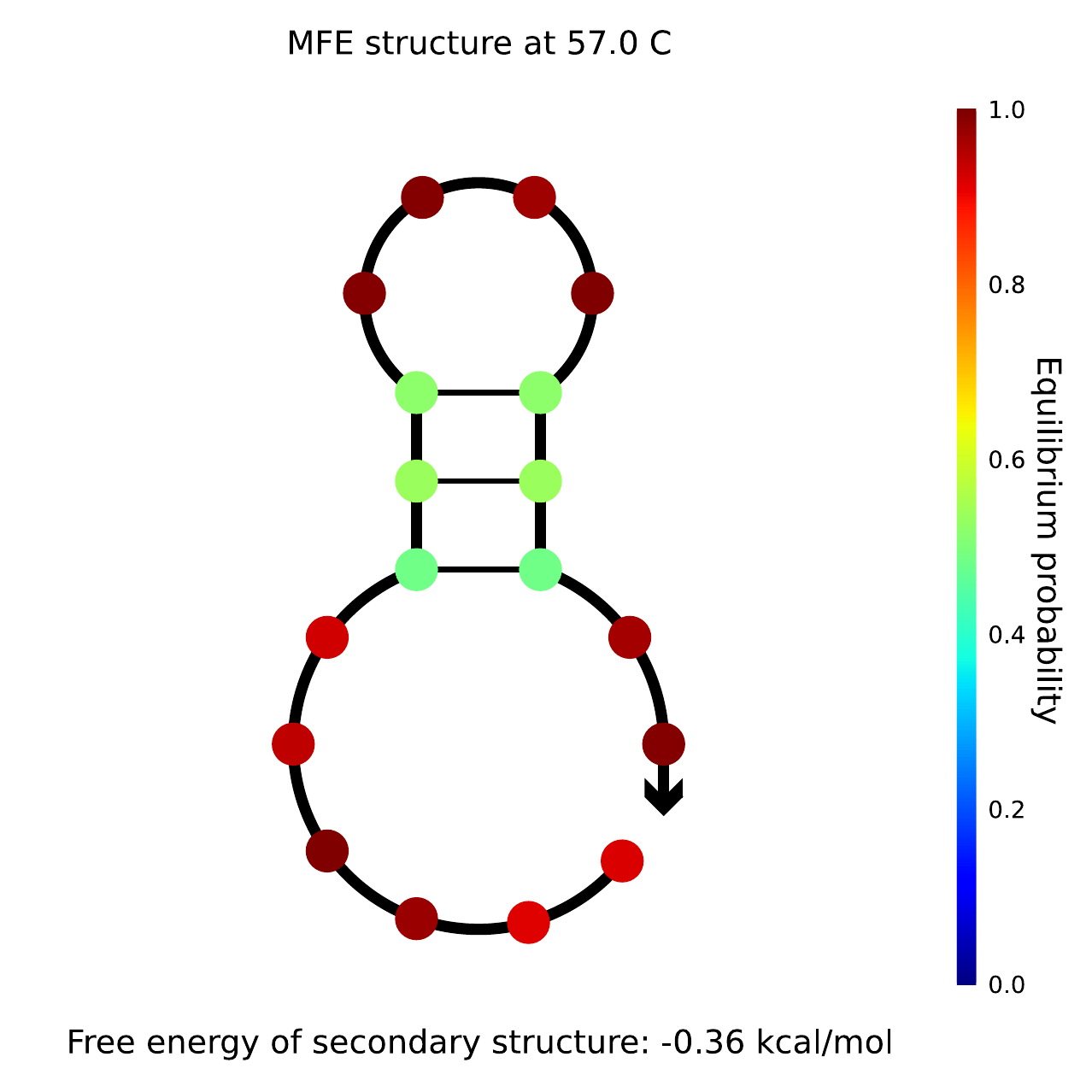}
    \caption{Equilibrium concentration of the ssDNA is only \SI{0.17}{\micro\mole}.}
    \label{fig:npkpairprob2}
  \end{subfigure}
  \hspace{60pt}
  \begin{subfigure}[t]{0.3\textwidth}
    \includegraphics[width=\textwidth]{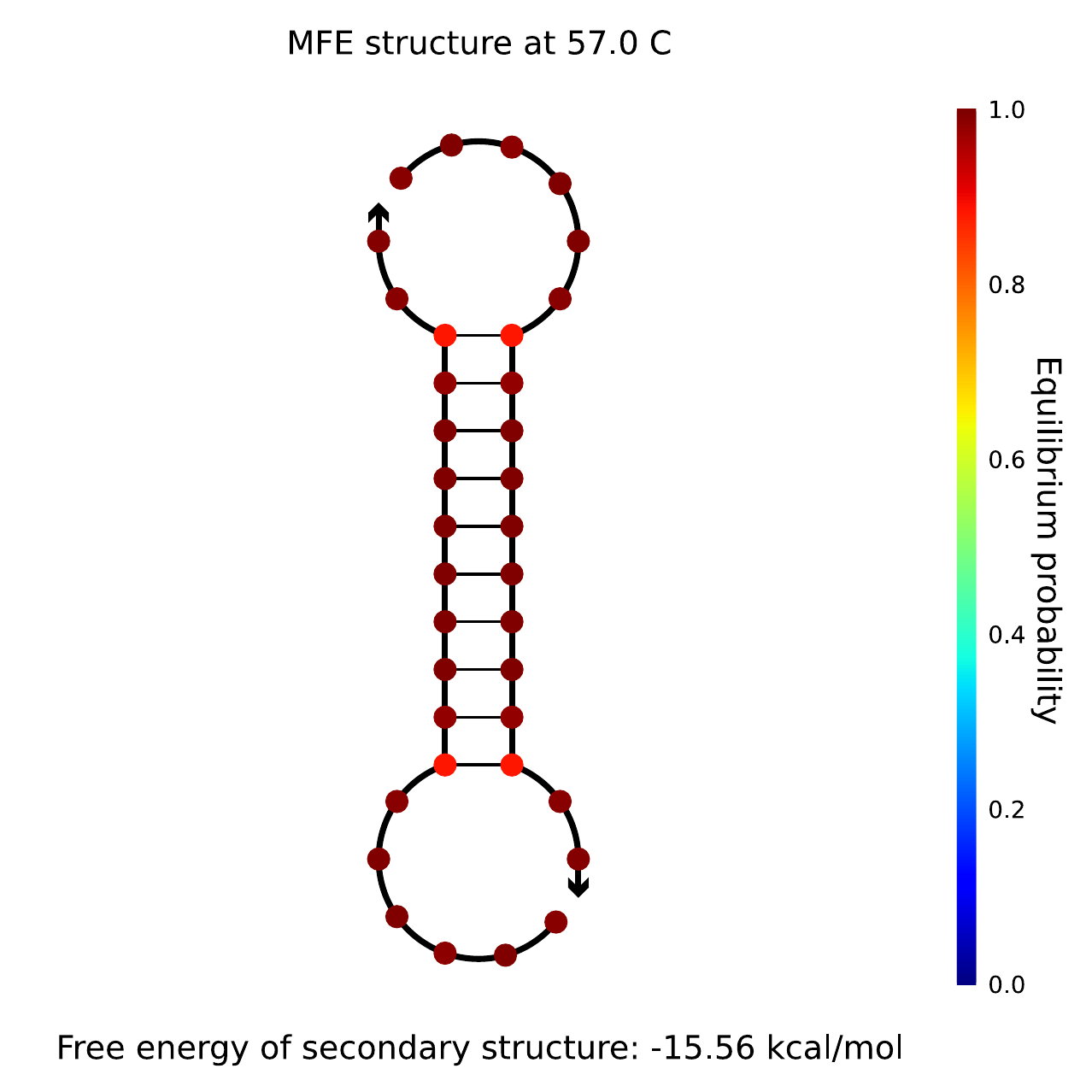}
    \caption{Equilibrium concentration of the dsDNA is \SI{0.92}{\micro\mole}.}
    \label{fig:npkpairprob1}
  \end{subfigure}
  \captionsetup{justification=justified, singlelinecheck=off}
  \caption{Alignment and thermodynamic properties for various DNA sequences as reported by NUPACK. \textbf{a}, \textbf{b}, Duplex structure predicted by NUPACK (see parasail alignments in \cref{apx:aln-scores}). The parasail alignment is not in full agreement with the predicted binding. \textbf{c}, \textbf{d}, For single-stranded DNA, a low MFE indicates a high probability that the molecule develops self-complementarity or knots. \textbf{c} A sequence that is expected to be stable (\texttt{AGTACAAGTAGGACAGGAAGATA}). \textbf{d}, A sequence that is expected to be more problematic in hybridisation reactions (\texttt{TTTCGCACGGACGAGGACGTCCGTTA}). \textbf{f}, \textbf{e}, A sequence can be similar enough to its reverse complement that it is more probable to find it in duplex formations with different instances of itself rather than in the normal single-stranded state. Illustrated is the sequence \texttt{CCATGGAGGCGCGCCTTT} in a complex of size 2, each strand initially present in solution at concentration \SI{1}{\micro\mole}. The duplex formation of this sequence (\textbf{f}) is more than 5 times as abundant as the single-stranded conformation (\textbf{e}).}
  \label{fig:npk_all}
\end{figure}

\noindent A simplified, schematic overview of the CNN architecture is provided in \cref{fig:cnn}. Our CNN implementation is described in \textit{Methods} and the precise architectural details and hyperparameters are discussed in \cref{tab:cnn_arch} and \cref{apx-sec:deep-learning-hyp}.
\\\\
By studying sequence-based algorithms, the existing knowledge from language processing can be transferred to DNA storage and computing. The string representation of DNA sequences is naturally fit for RNNs. Here, we propose a character-level RNN based on bi-directional Long Short-Term Memory (LSTM) \cite{10.1162/neco.1997.9.8.1735} layers.
\\\\
Compared to CNNs, which capture the spatial information within a receptive field and learn pixel-level filters, RNNs process the input data (characters) sequentially and maintain an internal memory state. Here, we hypothesise that the LSTM can effectively model the interaction between hybridising DNA strands, which requires capturing long-distance nucleobase relationships. Our choice of encoding DNA pairs also naturally suggests using a bi-directional implementation (see \cref{fig:rnn}). A simplified overview of the RNN architecture is provided in \cref{fig:rnn}, and the architectural details and hyperparameters are provided in \cref{apx-sec:deep-learning-hyp}.
\\\\
Finally, we turn our attention to the Transformer \cite{DBLP:journals/corr/VaswaniSPUJGKP17}, an architecture now prevalent in deep learning, with its origins in Natural Language Processing. We refer the readers to \cite{tay2020efficient} for an authoritative review, and to \cref{apx-bkg-transf} for a short summary. For this study, we adapt an established Transformer implementation, RoBERTa \cite{DBLP:journals/corr/abs-1907-11692}. Our use falls within the \textit{encoder-only} category: in language processing jargon, we formulate our task as a \emph{sentence-pair} regression problem, where the inputs are two \say{sentences} (DNA sequences) with the goal of outputting a numerical value (hybridisation yield). At a high level, the workflow is very similar to \cref{fig:rnn}, in the sense that the inputs are tokenised and propagated through multiple Transformer blocks. Distinctly from other models, RoBERTa requires two separate training phases: pre-training and fine-tuning. Our implementation is described in \textit{Methods}.

\subsection*{Deep learning accurately predicts hybridisation}
\begin{figure}[H]
  \centering
  \captionsetup{justification=centering}
  \begin{subfigure}[t]{0.49\textwidth}
    \centering
    \caption{}
    \includegraphics[width=1.0\textwidth]{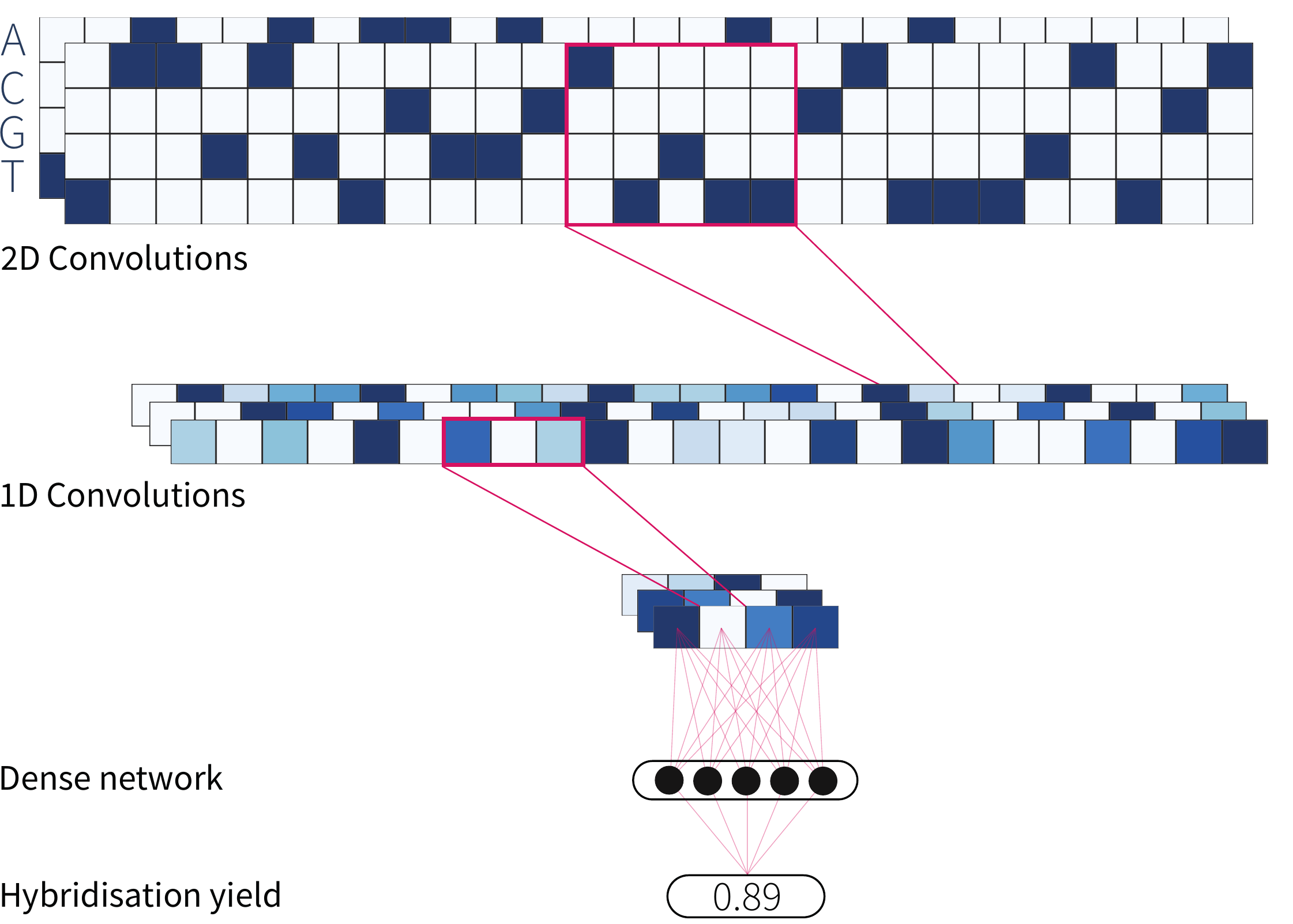}
    \label{fig:cnn}
  \end{subfigure}\hfill
  \begin{subfigure}[t]{0.49\textwidth}
    \centering
    \caption{}
    \includegraphics[width=1.0\textwidth]{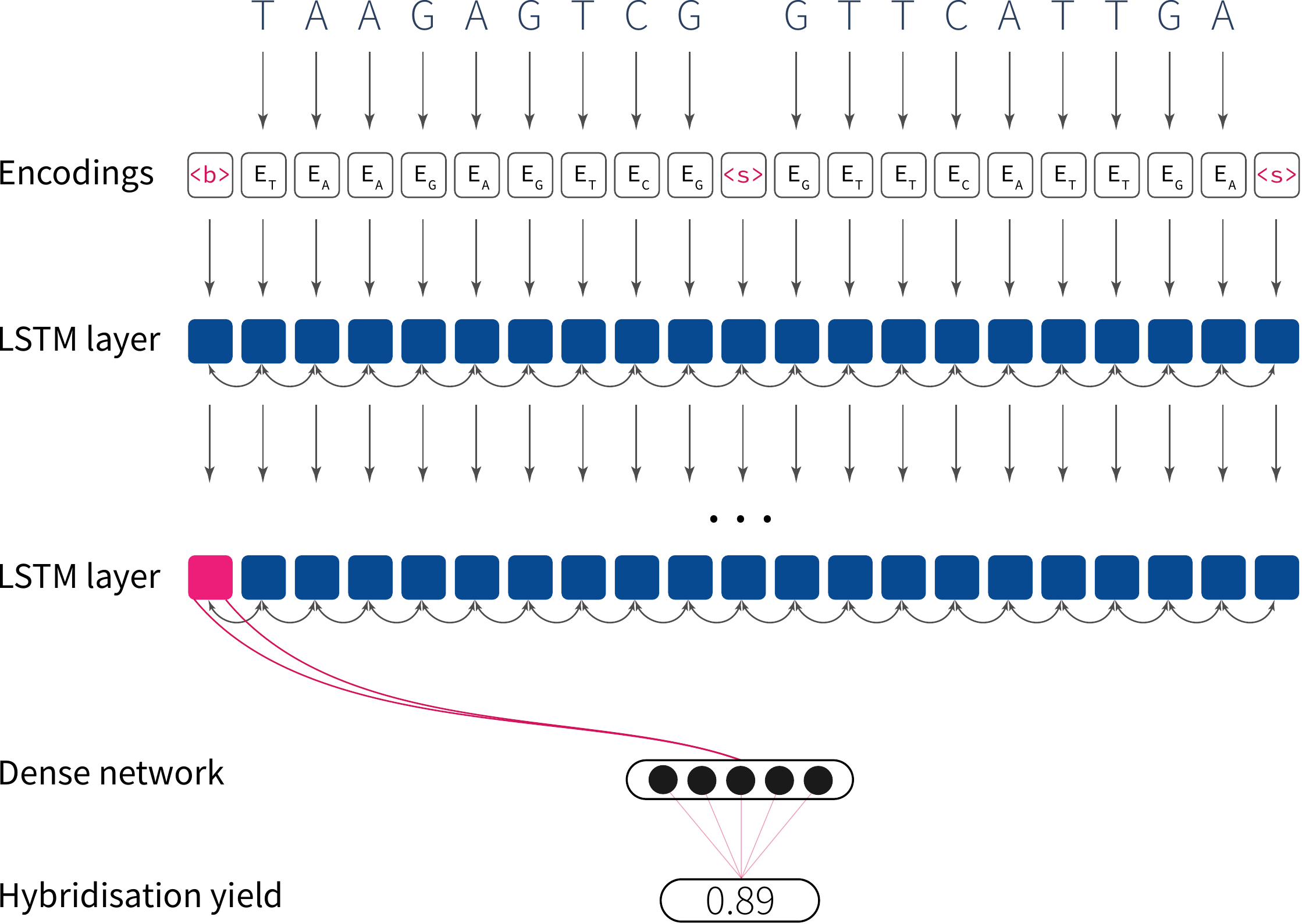}
    \label{fig:rnn}
  \end{subfigure}
  \begin{subfigure}[t]{0.49\textwidth}
    \centering
    \caption{}
    \includegraphics[width=1.0\textwidth]{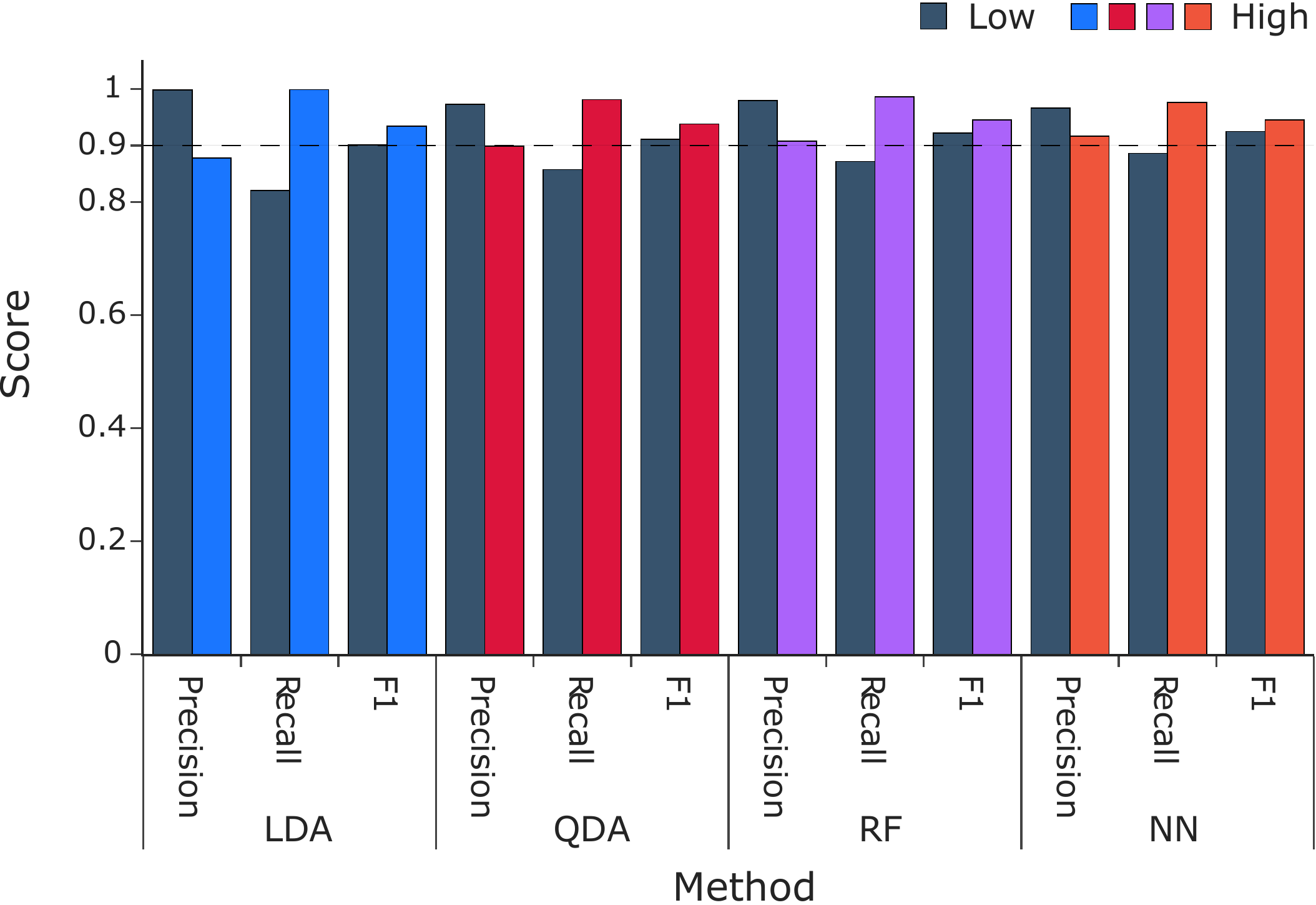}
    \label{fig:simple_ml_results}
  \end{subfigure}\hfill
  \begin{subfigure}[t]{0.49\textwidth}
    \centering
    \caption{}
    \includegraphics[width=1.0\textwidth]{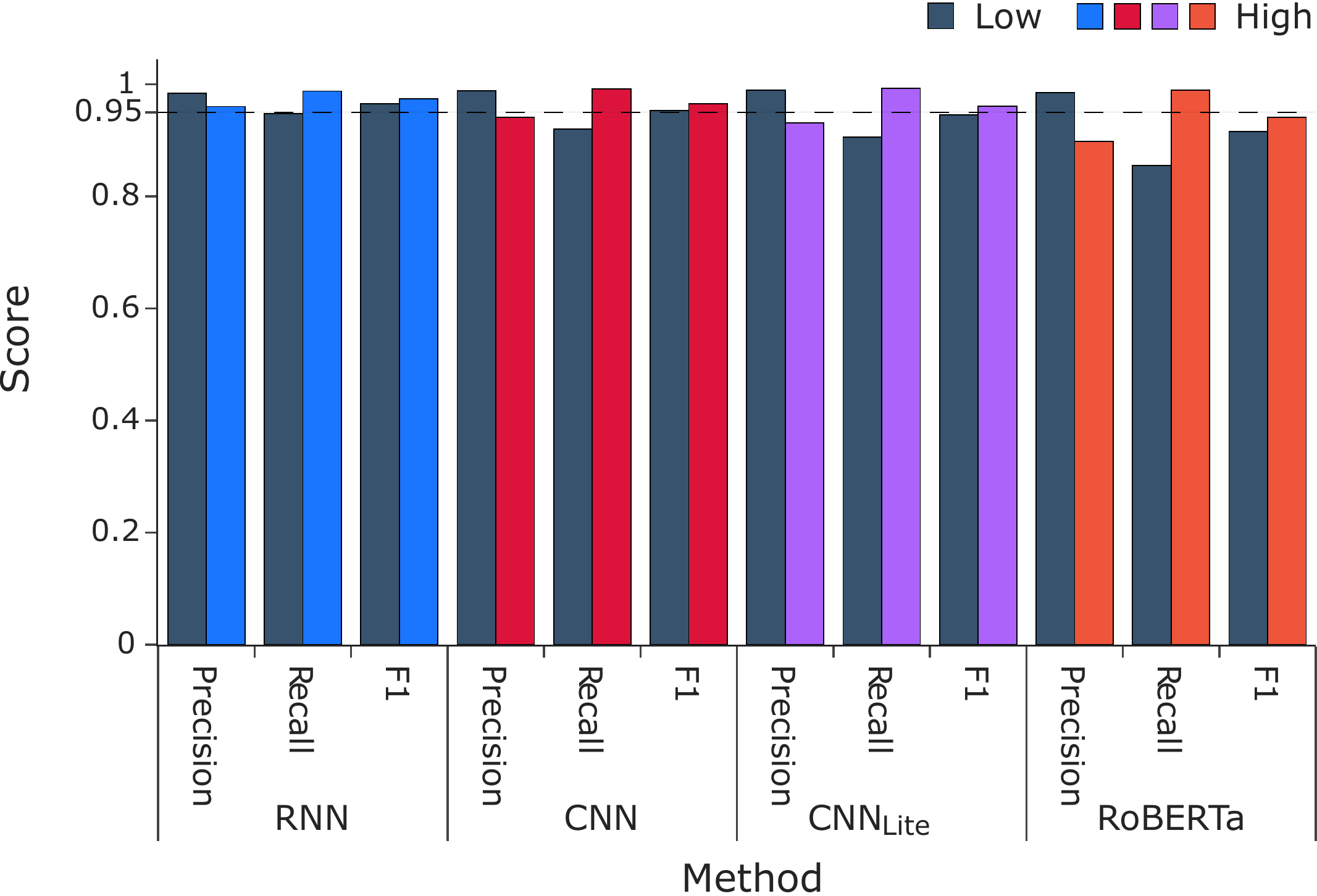}
    \label{fig:deep_learning_prec_rec_f1}
  \end{subfigure}
  \medskip
  \begin{subfigure}[b]{1.0\textwidth}
    \centering
    \caption{}
    \begin{tabular}{ccccc r@{.}l S[table-format=3.4]S[table-format=3.4]S[table-format=3.4]} \toprule
      Metric         & \textbf{LDA} & \textbf{QDA} & \textbf{RF} & \textbf{NN} & \multicolumn{2}{c}{\textbf{RNN}} & \textbf{\ \ CNN} & \textbf{$\text{CNN}_{\text{Lite}}$} & \textbf{RoBERTa}           \\ \midrule
      \textbf{MCC}   & 0.851        & 0.859        & 0.884       & 0.873       & \textbf{0}                       & \textbf{940}     & 0.920                               & 0.909            & 0.863   \\
      \textbf{AUROC} & 0.912        & 0.922        & 0.938       & 0.930       & \textbf{0}                       & \textbf{968}     & 0.956                               & 0.949            & 0.922   \\
      \textbf{MSE}   & N/A          & N/A          & N/A         & N/A         & \textbf{76}                      & \textbf{660}     & 109.550                             & 133.628          & 305.050 \\ \bottomrule
    \end{tabular}
    \label{tab:deep_learning_metrics}
  \end{subfigure}

  \captionsetup{justification=justified, singlelinecheck=off}
  \caption{\textbf{a}, \textbf{b}, Simplified representations of the deep learning architectures. \textbf{a}, Convolutional Neural Network overview. Each of the columns in the 2-channel grid (image) corresponds to a one-hot encoding of the four nucleobases for that particular strand position, while each channel represents an entire strand. 2D convolutions on the 2-channel one-hot encoded DNA strands are followed by 1D convolutions (only 3 channels shown) and fully-connected layers. \textbf{b}, Recurrent Neural Network overview. LSTM layers are widely used and recognised for their performance in language modelling tasks \cite{Irie2016}, as well as other sequence-based tasks \cite{chung2014empirical}. For readability, we represent bi-directional interactions with two-headed arrows between the sequence elements. \textbf{c}, \textbf{d}, \textbf{e}, Classification and regression results for the evaluated machine learning models. The choice of metrics is explained in \textit{Methods}. \textbf{c}, \textbf{d}, Graphical summary of precision, recall and $\text{F}_{\text{1}}$ score for the two classes of machine learning models. The \textbf{Low} class is represented by dark grey and the \textbf{High} class corresponds to the four bright colours (one colour for each method eases readability). \textbf{c}, The four baseline ML algorithms. Exact numerical values are provided in \cref{apx-tab:simple_ml_metrics}. \textbf{d}, Classification metrics for the four deep learning models, after yield binarisation (with a threshold of 0.2). The numerical values are provided in \cref{tab:deep_learnig_prec_rec_f1}. \textbf{e}, The AUROC and MCC summarise the classification performance of all evaluated machine learning techniques; additionally, the MSE (Mean Squared Error) is reported for the four deep learning models.}
  \label{fig:newfig}
\end{figure}

We evaluate three models termed simply CNN, RNN and RoBERTa and additionally a fourth convolutional model with a reduced number of convolutions and filter sizes: $\text{CNN}_{\text{Lite}}$. Performance metrics are summarised in \cref{fig:deep_learning_prec_rec_f1}, \ref{tab:deep_learning_metrics}.
With the exception of RoBERTa, we report a marked increase in classification performance over the baseline methods, which completely failed in approximating yields.
\\\\
We interpret the results at the level of false positives (FP) and false negatives (FN) as they enable a more fine-grained discussion. Examining the performance of the two best models (CNN and RNN), we find that the CNN model is slightly better at avoiding false negatives, at \num[group-separator={,},group-minimum-digits={3}]{1174} FNs compared to \num[group-separator={,},group-minimum-digits={3}]{1740} for the RNN. However, only 413 FNs are shared, meaning that a large portion of the FNs which are wrongly labelled by one model would be correctly classified by the other. It is important to note that the accurate labelling of the positives (\textbf{High} class) is an easier task than for negatives on our dataset, as indicated by the baseline methods.
\\\\
More interestingly, the RNN offers a pronounced decrease in false positives, at \num[group-separator={,},group-minimum-digits={3}]{5871} compared to \num[group-separator={,},group-minimum-digits={3}]{8981} for the CNN. In this case, \num[group-separator={,},group-minimum-digits={3}]{4732} FPs are shared, suggesting that, in general, the RNN makes the same mistakes as the CNN, but less often. False positives are the area where both the baseline and deep learning models struggled the most. FNs and FPs for the RNN indicate a trade-off, where the model loses some power in predicting true positives but significantly reduces false positives.
\\\\
The $\text{CNN}_{\text{Lite}}$ performs worse, on the whole, compared to the deeper CNN and the RNN. Generally, the erroneously classified samples are close to being subsets of the corresponding FNs and FPs of the CNN. While $\text{CNN}_{\text{Lite}}$ is not as strong a performer as the first two models, its benefits are in inference times, which will be presented shortly.

\setlength{\tabcolsep}{8pt}
\renewcommand{\arraystretch}{1.6}

\noindent The Transformer model, RoBERTa, has significant difficulties in discerning positives and exhibits the highest number of false positives, at \num[group-separator={,}]{16212}. About half of these are overlapping with those reported by the CNN model, and about a third are shared with the RNN. Thus, the dataset contains difficult examples that are not correctly classified by any model and this is exacerbated for RoBERTa. The number of false negatives is low, at \num[group-separator={,},group-minimum-digits={3}]{1174}, a trend observed by the other models as well. These mediocre results could be partially explained by Transformers being notoriously data-hungry. It is possible that not even our training dataset of over 2.5 million sequence pairs is enough to fully exploit the architecture.

\subsection*{Pre-trained models allow estimation on different temperatures}
\vspace{-4pt}
Studying the performance of the deep learning approach on different temperatures can prove or refute the generalisability of the models. Furthermore, for time and cost-saving purposes, it is essential to investigate if existing trained models can be repurposed for different temperature ranges that were not seen during training.
\\\\
For brevity, we report only the Matthews Correlation Coefficient in \cref{tab:mcc_other_temps}. Additional metrics are provided in \cref{tab:other_temps_extended} and \cref{tab:prec_rec_f1_other_temps}.
\\\\
We evaluate the models of \cref{fig:deep_learning_prec_rec_f1} on the five additional temperatures. \cref{tab:mcc_other_temps} and \cref{tab:other_temps_extended} suggest that the models perform reasonably well even on temperatures they were not trained on. The MSE continues to increase as the temperature is lowered and the distance from the training setting increases. For RNN, CNN and $\text{CNN}_{\text{Lite}}$, this also translates to a progressive alteration of the classification metrics. Overall, the RNN model is less generalisable, i.e., performs worse on different temperatures compared to the CNNs. The most dramatic drop in performance is for 62.0\textdegree{}, as high-scoring pairs tend to achieve much lower yields at this temperature.\\

\begin{table}[!t]
  \centering
  \captionsetup{justification=justified, singlelinecheck=off}
  \caption{MCC of the four deep learning models on the selected temperatures. The results suggest that the models perform reasonably well even on temperatures they were not trained on.}
  \begin{tabular}{cccccc} \toprule
    \textbf{Model}             & \textbf{37.0C} & \textbf{42.0C} & \textbf{47.0C} & \textbf{52.0C} & \textbf{62.0C} \\ \midrule
    RNN                        & 0.862          & 0.872          & 0.888          & 0.916          & 0.833          \\
    CNN                        & 0.884          & 0.893          & 0.908          & 0.929          & 0.811          \\
    $\text{CNN}_{\text{Lite}}$ & 0.893          & 0.901          & 0.914          & 0.929          & 0.801          \\
    RoBERTa                    & 0.925          & 0.930          & 0.930          & 0.914          & 0.762          \\ \bottomrule
  \end{tabular}
  \label{tab:mcc_other_temps}
\end{table}

\noindent Perhaps surprisingly, RoBERTa performs better in terms of classification metrics on the lower temperatures. In fact, this agrees with our previous observation that RoBERTa exhibits many false positives: as the temperature is lowered, low-scoring pairs are more likely to truly have high yields.

\subsection*{Deep learning reduces prediction time by orders of magnitude}
\vspace{-4pt}
Short inference times are crucial in scaling DNA hybridisation computations for the zettabyte future. To study this aspect, we perform a comprehensive empirical evaluation of the time required for the forward pass, i.e., prediction time. The choice of experimental platforms is described in detail in \cref{sec-app:exp_platform}, \cref{apx:software} and the approach taken to measure time in \textit{Methods} and \cref{sec-app:inf_time}. Results are reported in \cref{tab:inf_time}.
\\\\
Depending on the hardware used, we observe a speedup of one to over two orders of magnitude over multi-threaded NUPACK. When trained on a GPU, the previously identified best models (RNN and CNN) are around $\times23$ and respectively $\times16$ faster in making predictions, while TPUs almost trivialise inference times. The RNN's speed is comparable to the CNN's despite having less than one tenth of the number of trainable parameters. This is expected, as the sequential nature of RNNs prohibits the massively parallel optimisations possible for convolutions. For this reason, we introduced $\text{CNN}_{\text{Lite}}$, whose architecture is described in \cref{tab:cnn_lite_arch}. $\text{CNN}_{\text{Lite}}$ still offers competitive performance (\cref{tab:deep_learning_metrics}), while being almost twice as fast as the RNN. Due to the complex architecture and large number of parameters, RoBERTa (GPU) is slightly slower than NUPACK, while also underperforming in prediction accuracy. Given these two observations, it is difficult to currently recommend Transformers.
\\\\
The inference time increases proportionally with the dataset size (\Cref{apx:inf-diff-data-size}, \Cref{fig:apx-deep-learning-time}).

\begin{table}[t]
  \centering
  \captionsetup{justification=justified, singlelinecheck=off}
  \caption{The execution of all deep learning methods is timed on the test dataset of \num[group-separator={,}]{255701} samples. The average execution time and the standard deviation are reported in seconds. Each deep learning method is run 10 times, after an initial warm-up run. The time elapsed to load the dataset into memory is not taken into account and the batch size was chosen to maximise inference time. All deep learning models use consumer hardware or openly-available hardware (the TPU platform is completely free to use).}
  \begin{tabular}{cccc S[table-format=3.2] @{${}\hspace*{0.17cm}\pm{}$} S[table-format=2.2, table-number-alignment = left] S[table-format=-4.2]} \toprule
    \multicolumn{1}{c}{\textbf{Model}}          & \textbf{\# Params.}   & \textbf{Batch} & \multicolumn{1}{c}{\textbf{Hardware}} & \multicolumn{2}{c}{\textbf{Time (s)}} & \multicolumn{1}{c}{\textbf{Speedup}}                                           \\ \midrule
    NUPACK 3                                    & N/A                   & N/A            & 64-core VM                            & \multicolumn{2}{c}{372.59}            & $\times$1.00                                                                   \\ \addlinespace
    RoBERTa                                     & 6.1M                  & 1024           & RTX 3090                              & 388.44                                & 0.32                                 & $\times$0.96                            \\ \addlinespace
    \multirow{2}{*}{RNN}                        & \multirow{2}{*}{249K} & 8192           & RTX 3090                              & 15.87                                 & 0.10                                 & $\times$23.47                           \\
                                                &                       & 4096           & TPUv2                                 & 03.60                                 & 0.11                                 & $\times$103.50                          \\ \addlinespace
    \multirow{2}{*}{CNN}                        & \multirow{2}{*}{2.8M} & 512            & RTX 3090                              & 23.84                                 & 0.08                                 & $\times$15.63                           \\
                                                &                       & \textbf{4096}  & \textbf{TPUv2}                        & \hspace*{0.1cm}\textbf{01.23}         & \hspace*{0.12cm}\textbf{0.17}        & $\times$\hspace*{0.12cm}\textbf{301.74} \\ \addlinespace
    \multirow{2}{*}{$\text{CNN}_{\text{Lite}}$} & \multirow{2}{*}{470K} & 512            & RTX 3090                              & 09.01                                 & 0.00                                 & $\times$41.34                           \\
                                                &                       & 4096           & TPUv2                                 & 01.28                                 & 0.15                                 & $\times$290.21                          \\ \bottomrule
  \end{tabular}
  \label{tab:inf_time}
\end{table}

\subsection*{Integrating fast approximation into large-scale workflows}
We concentrate on the task of designing a set of orthogonal sequences. A brute-force approach to this problem quickly becomes unfeasible: checking every pair of a set is a quadratic problem in the length of the set: for a set of \num[group-separator={,}]{100000} initial sequences, over 5 billion pairs need to be checked. No matter how efficient the yield computation, we cannot tackle this problem using just brute-force. We can, however, devise a method that reduces the use of heuristics as much as possible and utilises our efficient yield approximation.
\\\\
Concretely, we determine an appropriate pairwise Longest Common Substring (not subsequence, i.e., the characters have to be consecutive) - the LCS. This is motivated by the fact that a sequence has to have at least \textit{some} similarity to its reverse complement for them to anneal. In particular, consecutive regions are much more stable than individual, separated nucleobases that match.
We demonstrate our workflow using a randomly generated 20-nucleobases sequence set of size \textbf{\num[group-separator={,}]{100000}}. We use MMseqs2 \cite{Steinegger2017MMseqs2Sets} to cluster sequences that achieve a minimum target LCS (in our case, 5). The full details are provided in \textit{Methods}. After processing with MMseqs2, we arrive at a set of \num[group-separator={,}]{20033335} pairs that passed the LCS threshold (i.e., have a consecutive overlap of at least 5 nucleobases) and that need to be further analysed by the yield computation.
\\\\
Compared to the brute-force approach, we have reduced the number of checks needed from over 5 billion to just $0.004\%$. Note that we do not lose anything in terms of accuracy, since the sequence pairs that we eliminate are so structurally different that it is meaningless to include them in the hybridisation computation.
\\\\
For NUPACK, we estimate a running time of over 62 hours -- two and a half days. Comparatively, based on our results from \cref{tab:inf_time} and \Cref{apx:inf-diff-data-size} we estimate that the RNN prediction time is around 6 minutes for second generation TPUs, with the CNNs expected to be even faster. This kind of workflow was previously impossible, and earlier strategies relied only on human-designed heuristics to filter sequences. With our approach, we apply precise estimations at a considerably larger scale than before.

\section*{Discussion}
We presented the first comprehensive study of machine learning techniques applied to the problem of predicting DNA hybridisation, aiming to improve DNA storage applications both qualitatively and quantitatively, and to encourage further research into this expanding area, especially by harnessing the power of machine learning for large, \textit{near-data processing}-enabled systems. As part of this investigation, we introduced a carefully designed dataset of more than 2.5 million sequence pairs with hybridisation scores for 6 different temperatures. We evaluated a range of classical machine learning methods as a baseline, and followed by introducing deep learning architectures capable of accurately predicting yields while reducing inference time by one to over two orders of magnitude. We believe that this combination of high-fidelity and massive throughput is necessary for scaling up emerging DNA storage applications. Our work can act as a building block for new neural architectures, or on their own, the deep learning models can be used for DNA database library design of a scope that was previously impossible. Future studies could focus on generative methods that produce pairs matching a given yield, predicting base-pair probabilities (i.e., at the character level) or applying Graph Neural Networks on DNA strands represented as graphs, for example for edge prediction (topology of the duplex).

%TC:ignore
\section*{Methods}
\subsection*{Dataset design}
We start with a small number (e.g. \num[group-separator={,},group-minimum-digits={3}]{1000} or \num[group-separator={,},group-minimum-digits={3}]{2000}) of randomly generated sequences where long repeats ($\geq 3$) of the same nucleobase are forbidden. We then perform three types of mutations: insertions, deletions and substitutions on each sequence. We incorporate both \textit{minor} mutations (generally limited to a single type of mutation out of the three and which is applied once or twice) and \textit{severe} mutations (all three types of mutations occur in the same sequence, possibly five or more times). To cover a wide range of thermodynamic interactions, we also apply the mutation operations on the reverse complement of the sequence. Thus, for each sequence, the mutation procedure generates a set of related sequences. We form a set of the initial randomly-generated sequences and all of their mutations, remove duplicates and generate all possible pairings. We repeat the procedure until the target size and variety constraints are achieved. To automate the yield computation process, we wrote Python wrappers for the NUPACK stand-alone executables. Each pair of inputs is written to a \texttt{.in} file, which is then fed to the \texttt{complexes} executable. In this work the default temperature is $57\degree$C, instead of the NUPACK default $37\degree$C; also see the discussion in \textit{Pre-trained models allow estimation on different temperatures}. As with the length criterion, this is based on the consensus in molecular biology that primers should be designed broadly in the $50\degree-65\degree$C range. The output of \texttt{complexes} is then used as input for the \texttt{concentrations} executable. Finally, we can read the result as a float and incorporate it into our dataset. Based on this numerical value, we can assign labels to each record. A good threshold, as suggested by \cite{Beliveau2018OligoMinerProbes}, is 0.2. Thus, pairs reported to be below 0.2 are labelled \textbf{Low}, otherwise \textbf{High}.

\subsection*{Alignment scores}
When used as a proxy for DNA annealing, we have to reverse complement one of the sequences. A high alignment score indicates that a sequence is very similar to the other's reverse complement. In this case, it is reasonable to assume that they will anneal. On the other hand, if the alignment score is low, then the sequence is dissimilar to the other's complement and we would not expect them to bind. However, as DNA interactions are very sensitive, it is possible to encounter pairs that do not strictly follow these rules. A few examples are illustrated in \cref{apx:aln-scores}, \cref{fig:npkaln2} and \cref{fig:npkaln1}. To perform alignments, we use parasail \cite{Daily2016}, a fast implementation of the Smith-Waterman and Needleman-Wunsch algorithms for local, and respectively global and semi-global sequence alignment. Out of the three alignment techniques, the one that fits our task is semi-global alignment. This is because we would expect that two compatible sequences can be aligned from end to end, perhaps with mismatches and gaps in-between. It is also possible that considerable ($>4$ nucleobases) deletions and insertions happened at the end of the sequences. This means that just global alignment is too weak, while local alignment is not optimal as we are not interested in matching isolated fragments of the sequences. The command used to compute the semi-global alignment of two sequences is:

\usemintedstyle{colorful}
\begin{minted}{python}
parasail.sg_scan_16(str(s1), str(s2), 5, 2, parasail.dnafull)
\end{minted}

\noindent where the input sequences \texttt{s1} and \texttt{s2} are Biopython \cite{10.1093/bioinformatics/btp163} \texttt{Seq} objects. Also note that we are using a DNA substitution matrix, a gap open penalty of 5 and a gap extend penalty of 2. The graphical representation of the alignment can be computed by enabling the traceback feature:
\usemintedstyle{colorful}
\begin{minted}{python}
parasail.sg_trace_scan_16(str(s1), str(s2), 5, 2, parasail.dnafull)
\end{minted}

\noindent The resulting alignments, however, are not always representative of the actual binding that NUPACK predicts, as can be seen when comparing \cref{apx:aln-scores} to the NUPACK outputs of \cref{fig:npkaln2} and \cref{fig:npkaln1}.

\subsection*{Classification metrics}
The overall classification performance can be expressed through two metrics, the Matthews Correlation Coefficient (MCC) and Area Under the Receiver Operating Characteristics (AUROC). The AUROC is well-known and widely used in classification tasks, while the MCC has been recently argued to be more informative and reliable in binary classification tasks compared to popular metrics like accuracy and $\text{F}_1$ score \cite{Chicco2020}, \cite{Chicco2017}. We do not reproduce the mathematical formulae here; however, they can be accessed through the cited studies.

\subsection*{Convolutional Neural Networks}
As exemplified in \cref{fig:npkaln2} and \cref{fig:npkaln1}, duplex interactions are complex and often do not correspond to a simple end-to-end alignment. To effectively incorporate this property, we used semi-local alignment for the baseline models. For CNNs, we first perform 2D convolutions with large filters of size $4 \times 9$, covering slightly under $35\%$ of the maximum sequence length. We follow with 1D convolutions of size $9$, $3$, $3$ and finally $1$. The large filters in the first two layers ensure that spatial information is captured from a wide area, while the following layers, decreasing in size, can learn more specialised features. Our experiments confirm this choice, as this architecture is superior to stacked convolutions of the same size, e.g. $3 \times 3$ (2D) then $3$ (1D) on our regression task. We also note that the small input size of $4 \times N \times 2$ limits the use of very deep models or alternative layers such as residual blocks.
\\\\
Similar encodings have been used in the literature, although not for pairs of sequences and mostly in different settings, for example in \cite{Stewart2018AEncodings}, \cite{Min2017}, \cite{Zhang2018}. In principle, an entire architecture can be devised using only 2D convolutions, by applying (zero) padding on the output of the first convolution operator (we used this approach in our work at DNA25). For this study, 1-dimensional convolutions follow the first 2D operation. This change saves millions of learnable parameters while reaching very similar performance. All convolutions are followed by ReLU activation and batch normalisation, with dropout interspersed throughout. Following the convolutional operations, fully-connected layers predict a single numerical value, the hybridisation yield. The entire network is trained under a supervised learning setting with NUPACK yields as ground truth.

\subsection*{Recurrent Neural Networks}
Our sequential pair encoding requires the concatenation of two single-stranded sequences, not before inserting a special \textit{separator} token (\texttt{<s>} in \cref{fig:rnn}). Additionally, a special \textit{beginning} token, represented by \texttt{<b>} in \cref{fig:rnn} is prepended to the resulting sequence. Symmetrically, the separator token also marks the ending (last character). Each character is indexed into a vocabulary and then fed through an embedding layer. At this point, multiple recurrent layers can be stacked. The hidden states of the special beginning token (\texttt{<s>}) are extracted from the last layer and ran through a dense layer that outputs the hybridisation yield. As LSTM layers do not possess the same interpretable hyperparameters as CNNs (e.g. filter sizes), we perform hyperparameter optimisation (see \cref{apx-sec:deep-learning-hyp}).

\subsection*{Transformers}
A short description of a general Transformer architecture and the standard pre-training schemes is provided in \cref{apx-bkg-transf}. We perform pre-training using the collection of all unique single-stranded DNA sequences in our dataset (totalling 2,579,555 samples). As we formulate our task at the character level, we have no appropriate equivalent for longer sequences of text and thus for the NSP task. Consequently, only MLM is used during pre-training, which we consider useful as predicting nucleobases based on the surrounding context and position is important for learning the structure of DNA sequences. The purpose of the pre-training scheme is learning to model DNA at the primary structure level. During fine-tuning, the pre-trained model is adjusted for the sentence-pair regression task by appending a fully-connected layer and propagating gradients throughout the entire model. We follow our intuition that modelling DNA interactions is not as complex as modelling natural language and since training Transformer models is extremely computationally demanding, we do not perform hyperparameter optimisation; instead, we base our architecture on documented models such as \texttt{roberta-base}. The particular framework we employ is HuggingFace's Transformers \cite{DBLP:journals/corr/abs-1910-03771}.

\subsection*{Measuring inference time}
For each platform, we perform 10 trials (repetitions) after one warm-up trial and report the results. We took all precautions we are aware of to ensure that measuring the inference time is as precise and fair as possible. NUPACK was executed on a 64-core Azure VM equipped with the premium SSD option (an important observation since NUPACK outputs must be written and read from disk). Since NUPACK is a single-threaded application, we wrote Python wrappers around NUPACK and employed a \texttt{ProcessPool} from the \texttt{pathos} multiprocessing library to distribute NUPACK logic, reads and writes to disk across all available cores. This dramatically improved NUPACK running time from more than 5 hours when running (in single-threaded mode) on an AMD Ryzen 5950X processor and a PCI Express 4.0 SSD to slightly more than 6 minutes on the Azure VM. This is a more than $\times50$ improvement over the out-of-the-box NUPACK performance. The execution of the code was timed using the \texttt{time} shell command preceding the script invocation and the wall clock time was read and reported.
\\\\
When measuring execution times on the GPU, additional measures were taken. More specifically, we overrode the PyTorch Lightning method \texttt{test\_step} to perform only the forward pass and no additional computations (such as the loss function), and surrounded the call to the \texttt{test} method of the PyTorch Lightning \texttt{Trainer} object with two \texttt{torch.cuda.Event(enable\_timing=True)} objects calling their \texttt{record} method. Before calculating the elapsed time, we wait for all kernels in all streams on the CUDA device to complete (\texttt{torch.cuda.synchronize}). A representative snippet of timing code for GPUs is provided in \cref{sec-app:inf_time}. For GPUs, the PyTorch \texttt{DataLoader} object does not use multiple workers (\texttt{num\_workers=0}). For consistency, we use the same setting for TPUs.
\\\\
Finally, to the best of our knowledge, there currently is no established way to measure execution time on Tensor Processing Units (TPUs). To combat this, we implemented our own timing code which is consistent with the epoch time reported by PyTorch Lightning (note that the epoch time only reports elapsed seconds, i.e., low temporal resolution). The PyTorch Lightning method \texttt{test\_step} is subjected to the same treatment as above. Our code defines a boolean flag inside the \texttt{LightningModule} to be timed; the first time the \texttt{forward} method is entered, the flag is set and the timestamp is recorded. The PyTorch Lightning method \texttt{on\_test\_epoch\_end} is also overridden to capture a final timestamp and compute the elapsed time. A representative snippet of timing code for TPUs is provided in \cref{sec-app:inf_time}.

\subsection*{Finding an appropriate Longest Common Substring}
We start by investigating our 2.5 million pairs dataset. In particular, we looked only at pairs with high yield (above 0.2), which correspond to about $56\%$ of the dataset, or more exactly, \num[group-separator={,},group-minimum-digits={3}]{1437758} elements. Out of the \num[group-separator={,},group-minimum-digits={3}]{1437758} high yield pairs, the shortest LCS among pairs is 4, which happens for only 29 pairs, or about $0.00002\%$ of the high yield set. Also, the 29 pairs have an average yield of only 0.37, which is overall still quite low, even though it passes the threshold of 0.2. The pairs with an LCS of 5 are more numerous, at 349 samples, with an average yield of 0.47. We have thus decided that for our next experiment, working with a \emph{minimum} LCS of 5 is the most appropriate. The running time performance difference between 4 and 5 is significant enough in the clustering stage described next to justify the trade-off. When working with sets of sequences of length 20, this means a minimum $25\%$ identity match between sequences, which we find reasonable. The problem is now to efficiently process a large dataset such that sequences with shared regions of $25\%$ or longer are associated. From our experience, the most appropriate tool for this task is the recently-developed MMseqs2. Briefly, MMseqs2 is described as an \say{\textit{ultra fast and sensitive search and clustering suite}}, and its performance is said to scale almost linearly with the number of processor cores.

\subsection*{Clustering with MMseqs2}
\label{sec-mth:mmseqs}
Assuming a set of single stranded DNA sequences in FASTA format, the first step is to build a MMseqs2 database, with the following command:

\usemintedstyle{borland}
\begin{minted}{bash}
mmseqs createdb <filename>.fasta <dbname> --dbtype 2 --max-seq-len <N>
\end{minted}

\noindent The \texttt{dbtype} option indicates a nucleotide database. For example, in practice the command might look like:

\usemintedstyle{borland}
\begin{minted}{bash}
mmseqs createdb init.fasta seqsDB --dbtype 2 --max-seq-len 20
\end{minted}

\noindent The crucial step is the search. A fully configured search command looks like the following:

\usemintedstyle{borland}
\begin{minted}{bash}
mmseqs search <dbname> <dbname> <alignname> <tmp> -s 7.5 -k 5
--max-seqs 10000 --exact-kmer-matching 1 --spaced-kmer-mode 0
--min-seq-id 0.25 --alignment-mode 4 -e inf
--min-aln-len 5 --sub-mat dnafull --max-seq-len 20
--strand 0 --search-type 3
\end{minted}

\noindent A short description is in order. The \texttt{dbname} is the same as used in the database creation step, while \texttt{alignname} and \texttt{tmp} are names for the output. The sensitivity is set to the maximum (7.5). The search is to be done using $k$-mers of size 5. We allow a maximum of \num[group-separator={,},group-minimum-digits={3}]{10000} related sequences to be returned for each individual query sequence. Exact $k$-mer matching is enabled, and spaced $k$-mer mode is disabled to ensure alignment only of ungapped regions. The minimum sequence similarity is defined to be 0.25. Alignment mode 4 is faster, and only considers ungapped alignments. The \texttt{-e inf} option targets similarity, and we want to consider all possibilities. The minimum alignment length is set to 5. The substitution matrix is set to \texttt{dnafull}, the one used for DNA sequences. We again specify a maximum sequence length of 20. The option \texttt{--strand 0} is important, as it only searches for similar regions in the reverse complements of the set. Finally, a nucleotide search is specified by \texttt{--search-type 3}.
\\[12pt]
The low running time of this search allows this procedure to be carried out even on a laptop. On our portable experimental platform with 12 CPU threads, the search takes under 15 minutes. We have also verified the search speed on an Azure VM with 64 CPUs and the search finishes in under 5 seconds. The generated output file has \num[group-separator={,},group-minimum-digits={3}]{39432713} pairs, with a small sample reproduced in \cref{tab:mmseqsres}.
\\\\
We notice that it is possible for some sequences to be associated with themselves, i.e., have a pair of identical IDs. We eliminate these from the set, discovering that there are just \num[group-separator={,},group-minimum-digits={3}]{7056} such pairs. We also notice that it is possible for symmetric pairs to occur, i.e., a pair such as \texttt{(seq10, seq20)} and \texttt{(seq20, seq10)}. By eliminating one of these duplicates, we are left with \num[group-separator={,},group-minimum-digits={3}]{20033335} pairs out of the \num[group-separator={,},group-minimum-digits={3}]{39425657} from the previous filtering step, which is a $51\%$ smaller set.

%TC:endignore
\section*{Data Availability}
The author declares that the dataset and the source code supporting the findings of this study are hosted on GitHub at https://github.com/davidbuterez/dna-hyb-deep-learning.

\section*{Author contributions statements}
D.B. conceptualised the study, ran the experiments, wrote the manuscript text and prepared all the figures.

\section*{Competing interests}
The author declares no competing interests.

\clearpage
\printbibliography

\resetlinenumber
%TC:ignore
\appendixpageoff
\appendixtitleoff
\renewcommand{\appendixtocname}{Supplementary information}
\renewcommand{\figurename}{Supplementary Figure}
\renewcommand{\tablename}{Supplementary Table}
\crefalias{figure}{appendixfigure}
\crefalias{subfigure}{appendixfigure}
\crefalias{table}{appendixtable}
\crefalias{section}{appendixname}
\setcounter{figure}{0}
\setcounter{table}{0}
\clearpage

\clearpage
\begin{appendices}

  \setcounter{page}{1}
  \chapter*{Supplementary information}
  \vspace{-40pt}

  \section{Alignment scores}
  \label{apx:aln-scores}
  \noindent Notice that there are discrepancies between the parasail alignments below and the structures reported by NUPACK in \cref{fig:npkaln2} and \cref{fig:npkaln1}.

  \setlength\columnsep{90pt}
  \begin{multicols}{2}
    \begin{itemize}[wide=0pt]

      \item[] \textbf{Score} $130$ - \textbf{Yield} $1.0$:
            \begin{verbatim}
 GAATACTGTCAGTGAGAGGATCTGCC
 ||||||||||||||||||||||||||
 GAATACTGTCAGTGAGAGGATCTGCC
  \end{verbatim}

      \item[] \textbf{Score} $77$ - \textbf{Yield} 0.12:
            \begin{verbatim}
 TTGTCATACGCTGTAAGAG
 |||||||.|||||||||.|
 TTGTCATCCGCTGTAAGCG
  \end{verbatim}

            \columnbreak

      \item[] \textbf{Score} $31$ - \textbf{Yield} 0.20:
            \begin{verbatim}
 ----C-TGCGGCGCCGTTTGCATGCTCTCG
     | ..|||||||    ||| |
 AGAGCAAACGGCGCC----GCA-G------
  \end{verbatim}

      \item[] \textbf{Score} $86$ - \textbf{Yield} 0.82:
            \begin{verbatim}
 AAATCAGGTA---TGCGGTAAG
 ||||||||||   |||||||||
 AAATCAGGTACGTTGCGGTAAG
  \end{verbatim}

    \end{itemize}
  \end{multicols}

  \section{Hyperparameters for LDA, QDA, RF and NN}
  \label{apx-sec:simple_ml_hyperparam}
  The hyperparameter optimisation for the baseline machine learning algorithms is performed using sklearn \cite{Pedregosa2011Scikit-learn:Perrot}, more specifically with the \texttt{GridSearchCV} class configured to use our validation split (possible using \texttt{PredefinedSplit}). The monitored metric is the Matthews Correlation Coefficient, as advised in \cite{Chicco2020}, \cite{Chicco2017}. The chosen search space is inspired by the default hyperparameter settings and is meant to provide reasonable coverage, keeping in mind a trade-off between exhaustiveness and time.

  \subsubsection*{LDA}
  \vspace{-4pt}
  For LDA, the number of hyperparameters is small. We use a parameter grid with two entries:
  \usemintedstyle{bw}
  \begin{minted}{python}
{solver: [lsqr, eigen], shrinkage: [auto, *np.linspace(0.0, 1.0, num=11)]},
{solver: [svd]}
\end{minted}
  The found (and used) hyperparameters are: \mintinline{python}|solver: eigen, shrinkage: 0|.

  \subsubsection*{Random forests}
  \vspace{-4pt}
  For random forests, the search space included:
  \usemintedstyle{bw}
  \begin{minted}{python}
{n_estimators: [30, 100, 200], max_depth: [None, 10, 30, 100],
max_features = [None, auto, sqrt], min_samples_split = [2, 5],
min_samples_leaf = [1, 2, 4], bootstrap = [True, False]}
\end{minted}
  The returned hyperparameters were \mintinline{python}|bootstrap: True, max_depth: 10, max_features: sqrt|\\
  \mintinline{python}|min_samples_leaf: 1, min_samples_split: 2, n_estimators: 100|. However, training with default parameters resulted in better performance on the test set. Thus, the reported results make use of the default hyperparameters.

  \subsubsection*{Neural networks}
  \vspace{-4pt}
  In the case of neural networks we tuned hyperparameters using Optuna \cite{optuna2019}, an open-source optimisation framework that includes state-of-the-art algorithms. In particular, Optuna also includes a pruning capability to stop training unpromising trials. The hyperparameters included in the search are: the number of hidden dense layers $\texttt{n\_layers} \in \{1, 2, 3, 4, 5\}$, a single dropout value to be applied after every hidden layer $\texttt{dropout} \in [0.2, 0.5]$, the number of neurons per hidden layer $\texttt{n\_units} \in \{4, 5, ..., 128\}$, the batch size $\texttt{batch} \in \{32, 64, 128, 256, 512\}$ and the learning rate $\texttt{lr} \in [0.0001, 0.1]$ (log domain). The neural optimisation algorithm is Adam \cite{kingma2017adam}.
  \\\\
  Optuna was run with the \texttt{HyperbandPruner} for 50 trials with the goal of minimising the validation loss (binary cross-entropy as it is a classification task). The returned (and used) best hyperparameters were: \mintinline{python}|n_layers: 4, dropout: 0.3296, n_units_l0: 117,| \mintinline{python}|n_units_l1: 18, n_units_l2: 7,| \mintinline{python}|n_units_l3: 19, batch: 1024, lr: 0.0002| (rounded to 4 decimal places).

  \section{Classification metrics for LDA, QDA, RF and NN}
  \label{apx-sec-simple-ml}
  The following table provides the basis for our discussion on the classification performance of the baseline ML models.

  \vspace{-5pt}
  \setlength{\tabcolsep}{6pt}
  \begin{table}[H]
    \centering
    \caption{Precision, recall and $\text{F}_{1}$ corresponding to \cref{fig:simple_ml_results}.}
    \vspace{6pt}
    \label{apx-tab:simple_ml_metrics}
    \begin{tabular}{ccccccccc} \toprule
                                       & \multicolumn{2}{c}{\textbf{LDA}} & \multicolumn{2}{c}{\textbf{QDA}} & \multicolumn{2}{c}{\textbf{RF}} & \multicolumn{2}{c}{\textbf{NN}}                                                               \\ \midrule
      \textbf{Metric}                  & \textbf{Low}                     & \textbf{High}                    & \textbf{Low}                    & \textbf{High}                   & \textbf{Low} & \textbf{High} & \textbf{Low} & \textbf{High} \\ \cmidrule(lr){1-1} \cmidrule(lr){2-3} \cmidrule(lr){4-5} \cmidrule(lr){6-7} \cmidrule(lr){8-9}
      \textbf{Precision}               & 0.996                            & 0.881                            & 0.975                           & 0.901                           & 0.966        & 0.927         & 0.971        & 0.914         \\
      \textbf{Recall}                  & 0.827                            & 0.998                            & 0.860                           & 0.983                           & 0.901        & 0.975         & 0.881        & 0.980         \\
      $\textbf{\text{F}}_{\mathbf{1}}$ & 0.904                            & 0.936                            & 0.914                           & 0.940                           & 0.932        & 0.950         & 0.924        & 0.946         \\ \bottomrule
    \end{tabular}
  \end{table}

  \section{Classification metrics for CNN, RNN and RoBERTa}
  \label{apx-sec-deep-learning}
  Similarly to the above, we report the classification metrics for the deep learning models after applying the 0.2 threshold on the predicted yields (binarisation).

  \vspace{-5pt}
  \setlength{\tabcolsep}{6pt}
  \begin{table}[H]
    \centering
    \caption{Precision, recall and $\text{F}_{1}$ corresponding to \cref{fig:deep_learning_prec_rec_f1}.}
    \vspace{6pt}
    \label{tab:deep_learnig_prec_rec_f1}
    \begin{tabular}{ccccccccc} \toprule
                                       & \multicolumn{2}{c}{\textbf{RNN}} & \multicolumn{2}{c}{\textbf{CNN}} & \multicolumn{2}{c}{\textbf{{$\text{CNN}_{\text{Lite}}$}}} & \multicolumn{2}{c}{\textbf{RoBERTa}}                                                               \\ \midrule
      \textbf{Metric}                  & \textbf{Low}                     & \textbf{High}                    & \textbf{Low}                                              & \textbf{High}                        & \textbf{Low} & \textbf{High} & \textbf{Low} & \textbf{High} \\ \cmidrule(lr){1-1} \cmidrule(lr){2-3} \cmidrule(lr){4-5} \cmidrule(lr){6-7} \cmidrule(lr){8-9}
      \textbf{Precision}               & 0.984                            & 0.960                            & 0.989                                                     & 0.941                                & 0.990        & 0.931         & 0.985        & 0.898         \\
      \textbf{Recall}                  & 0.948                            & 0.988                            & 0.920                                                     & 0.992                                & 0.906        & 0.993         & 0.855        & 0.990         \\
      $\textbf{\text{F}}_{\mathbf{1}}$ & 0.965                            & 0.974                            & 0.953                                                     & 0.966                                & 0.946        & 0.961         & 0.915        & 0.942         \\ \bottomrule
    \end{tabular}
  \end{table}

  \clearpage
  \section{Hyperparameters for CNN, RNN and RoBERTa}
  \label{apx-sec:deep-learning-hyp}
  As the three architectures have innate differences, our approach to designing each model is different.

  \subsubsection{CNN}
  \vspace{-4pt}
  For our CNN model, we follow a top-down approach based on the properties of our inputs (pairs of DNA sequences represented as grids). At a high level, the network layers can be grouped in convolutional blocks, where each block has a convolutional layer, an activation (ReLU) layer and a batch normalisation layer (in this order), with dropout layers interspersed according to \cref{tab:cnn_arch}. The model is trained with the Adam optimiser and a learning rate of 0.0001, a batch size of 256 (maximum that would fit in 8GB of GPU memory) with the Minimum Squared Error (MSE) loss and early stopping set to a patience of 3 epochs and no maximum number of epochs.
  \\\\
  We also performed hyperparameters search for the CNN architecture. The search space was defined by: number and size of filters $\texttt{conv2d\_size} \in \{9, 10, 11, 12, 13\}$, $\texttt{conv1d\_size1} \in \{7, 8, 9, 10, 11\}$, \\ $\texttt{conv1d\_size2}, \texttt{conv1d\_size3} \in \{3, 4, 5, 6, 7\}$, $\texttt{conv1d\_size4} \in \{1, 2, 3\}$, $\texttt{conv2d\_filter},\\ \texttt{conv1d\_filter1}, \texttt{conv1d\_filter2} \in \{256, ..., 768\}$, $\texttt{conv1d\_filter3} \in \{128, ..., 384\}$,\\ $\texttt{conv1d\_filter4} \in \{32, ..., 128\}$, the batch size $\texttt{batch\_size} \in \{ 256, 512, 1024, 2048, 4096, 8096\}$ and the learning rate $\texttt{lr} \in [0.0001, 0.1]$ (log domain). Dropout can be applied after each convolutional layer, with a choice of probability in $[0, 0.5]$ and after the linear layers with probability in $[0.1, 0.5]$. The choice of size for the linear layers is based on the value selected for $\texttt{conv1d\_filter4}$. The values found by the hyperparameter search are available in the repository. However, they were not used as the classification performance on the test set using the reported hyperparameters was slightly worse than the architecture described in \cref{tab:cnn_arch}.

  \begin{table}[h]
    \centering
    \caption{Overview of the CNN architecture.}
    \label{tab:cnn_arch}
    \begin{tabular}{lccc} \toprule
      \multicolumn{1}{c}{\textbf{Layer}} & \textbf{In channels} & \textbf{Out channels} & \textbf{Filter or dropout} \\ \midrule
      2D Convolution                     & 2                    & 512                   & $4 \times 9$               \\
      Dropout                            & *                    & *                     & 0.2                        \\
      1D Convolution                     & 512                  & 512                   & 9                          \\
      1D Convolution                     & 512                  & 128                   & 3                          \\
      Dropout                            & *                    & *                     & 0.2                        \\
      1D Convolution                     & 128                  & 128                   & 3                          \\
      1D Convolution                     & 128                  & 64                    & 1                          \\
      Fully connected                    & 384                  & 256                   & *                          \\
      Fully connected                    & 256                  & 128                   & *                          \\
      Dropout                            & *                    & *                     & 0.2                        \\
      Fully connected                    & 128                  & 1                     & *                          \\ \bottomrule
    \end{tabular}
  \end{table}

  \begin{table}[h]
    \centering
    \caption{Overview of the $\text{CNN}_{\text{Lite}}$ architecture.}
    \label{tab:cnn_lite_arch}
    \begin{tabular}{lccc} \toprule
      \multicolumn{1}{c}{\textbf{Layer}} & \textbf{In channels} & \textbf{Out channels} & \textbf{Filter or dropout} \\ \midrule
      2D Convolution                     & 2                    & 256                   & $4 \times 9$               \\
      Dropout                            & *                    & *                     & 0.2                        \\
      1D Convolution                     & 256                  & 128                   & 9                          \\
      1D Convolution                     & 128                  & 64                    & 3                          \\
      Fully connected                    & 512                  & 256                   & *                          \\
      Dropout                            & *                    & *                     & 0.2                        \\
      Fully connected                    & 256                  & 1                     & *                          \\ \bottomrule
    \end{tabular}
  \end{table}

  \subsubsection*{RNN}
  \vspace{-4pt}
  The RNN architecture includes an embedding layer, a configurable multi-layer bi-directional LSTM block, a dropout layer and the fully-connected regression layer (in this order). The parameters of the LSTM-based model are not as intuitive as their CNN counterpart and we address this issue by performing hyperparameter optimisation using Optuna. The search space includes: the embedding dimension $\texttt{emb\_dim} \in \{ 16, 32, 64 \}$, the number of LSTM layers $\texttt{n\_layers} \in \{ 1, 2, 3, 4 \}$, the LSTM dropout $\texttt{lstm\_dropout} \in [0.1, 0.5]$, the number of features in the hidden state $\texttt{hidden} \in \{ 32, 64, 100 \}$, the dropout preceding the regression layer $\texttt{lin\_dropout} \in [0.1, 0.5]$, the batch size $\texttt{batch} \in \{ 64, 128, 256 \}$ and the learning rate $\texttt{lr} \in [0.0001, 0.1]$ (log domain).
  As before, the model is trained with the MSE loss and early stopping set to a patience of 3 epochs and no maximum number of epochs.
  \\\\
  The returned (and used) RNN hyperparameters are \mintinline{python}|emb_dim: 32, lstm_dropout: 0.2412|\\ \mintinline{python}|n_layers: 3, hidden: 64, lin_dropout: 0.2166, batch: 128, lr: 0.006| (rounded to 4 decimal places).
  \subsubsection*{RoBERTa}
  Transformer models require substantially more computing power to train end-to-end (including both the pre-training and the fine-tuning stages). As such, it is not feasible to perform hyperparameter optimisation for RoBERTa. However, we base our configuration on known RoBERTa architectures such as \texttt{roberta-base} and on the intuition that modelling DNA sequence hybridisation is an easier task than natural language processing.
  \\\\
  As described in the main text, the Transformer model is first pre-trained on a Masked Language Model (MLM) task after the inputs have been tokenised with \texttt{RobertaTokenizerFast} (the provided tokenisation class). The pre-training phase uses the following hyperparameters: vocabulary size $\texttt{vocab\_size} = 5000$, dimension of the encoder and pooler layers $\texttt{hidden\_size} = 256$, number of encoder hidden layers $\texttt{num\_hidden\_layers} = 6$, number of attention heads for each attention layer of the encoder $\texttt{num\_attention\_heads} = 8$, dimensionality of the intermediate layer $\texttt{intermediate\_size} = 1024$, the maximum sequence length that the model might ever use $\texttt{max\_position\_embeddings} = 128$, vocabulary setting $\texttt{type\_vocab\_size} = 1$, dropout probability for all fully connected layers in the embeddings $\texttt{hidden\_dropout\_prob} = 0.3$ (also in the encoder and pooler), dropout ratio of attention probabilities $\texttt{attention\_probs\_dropout\_prob} = 0.3$, Masked Language Model masking probability $\texttt{mlm\_probability}=0.15$, number of warm-up steps $\texttt{warmup\_steps}=500$, number of per-device batch size $\texttt{per\_device\_train\_batch\_size}=256$ and the total number of pre-training epochs $\texttt{num\_train\_epochs}=6$.
  \\\\
  The other parameters are left to the default values. Our higher than default dropout values are used to help combat potential overfitting in the case where the MLM training task is too easy on our sequences. We find that around 4-6 epochs are enough for the model to converge on the pre-training phase.
  \\\\
  For the fine-tuning phase, we load the entire pre-trained RoBERTa model with trainable weights and append a dropout layer with probability $0.1$ and a dense layer for regression of the same size as the encoder layers (256). The input preprocessing employs the pair encoding capabilities of the tokeniser (initially designed for sentence pair tasks) and is very similar to the RNN encoding with special characters from \cref{fig:rnn}. The resulting network is trained with the MSE loss, AdamW optimiser with a learning rate of 0.00002 and weight decay set to 0.01, with early stopping set to a patience of 3 epochs and no maximum number of epochs.

  \section{Background on Transformers}
  \label{apx-bkg-transf}
  The term Transformer usually translates to a multi-layer architecture built from composable, structurally-identical Transformer \emph{blocks}, the fundamental units. The input to such a block is a $d$-dimensional vector, the additive composition of the input embedding vector itself with a separately-computed positional embedding vector. An important characteristic of the Transformer block is the self-attention mechanism, which computes an $N \times N$ attention matrix, with $N$ the sequence length. This procedure allows tokens to attend to other, possibly distant tokens which are deemed relevant (i.e., high attention scores). Having $k$-head attention corresponds to $k$ independent and concurrent attention computations that can be concatenated or summed. As noted in \cite{tay2020efficient}, Transformer blocks can be employed for different purposes. The three classic cases are \textit{encoder-only}, \textit{decoder-only} and \textit{encoder-decoder}, where each comes with specific implementation decisions and design choices. Naturally, the above computation is quadratic, and many recent variations have been proposed to combat this lack of efficiency \cite{tay2020efficient}.
  \\\\
  The original pre-training procedure includes two training objectives: Masked Language Model (MLM) and Next Sentence Prediction (NSP). MLM randomly replaces a per cent of input tokens with a special \textit{mask} token and learns to predict the missing character. NSP is a classification problem where the objective is to predict if two sentences follow each other in the input text.
  \\\\
  A possible extension is the \textit{Vision Transformer} \cite{dosovitskiy2020image}, a Transformer architecture designed to work on images, offering competitive performance with deep CNNs while requiring less resources. We do not investigate Vision Transformers in this work as the inference times will likely lag behind our relatively shallow CNNs, while the regression performance is also unlikely to be stellar considering our results with the sequence-based Transformer.

  \section{Clustering with MMseqs2}
  \label{apx:mmseqs2}

  A small sample of the \num[group-separator={,},group-minimum-digits={3}]{39432713} output pairs is reproduced in \cref{tab:mmseqsres}. Notice that for each ID in the first column, there are multiple sequences which are deemed similar by MMSeqs2. For example, the sequence denoted by ID \texttt{seq43840} has 8 other similar sequences associated with it. This is however just a subset of all the sequences found to be similar with \texttt{seq43840} and reproduced here for illustration purposes.

  \begin{table}[h]
    \centering
    \caption{Small sample of the MMseqs2 output.}
    \begin{tabular}{cccc} \toprule
      \textbf{ID 1} & \textbf{ID 2} & \textbf{Sequence 1}           & \textbf{Sequence 2}           \\ \midrule
      seq43840      & seq49789      & \texttt{GCGCCACCGCGTATATTAGG} & \texttt{AAGCTTAATACACGCGGTGC} \\
      seq43840      & seq81414      & \texttt{GCGCCACCGCGTATATTAGG} & \texttt{ATACGCGGTGGATGCGTAGC} \\
      seq43840      & seq92992      & \texttt{GCGCCACCGCGTATATTAGG} & \texttt{CATACGCGGCGGCGTCATAA} \\
      seq43840      & seq11505      & \texttt{GCGCCACCGCGTATATTAGG} & \texttt{GTTCTAATCTACGCGGAGTC} \\
      seq43840      & seq82545      & \texttt{GCGCCACCGCGTATATTAGG} & \texttt{GCACTTCATATATGCGGTGG} \\
      seq43840      & seq19444      & \texttt{GCGCCACCGCGTATATTAGG} & \texttt{ATACGCCGTGGCGGCCGGCA} \\
      seq43840      & seq15232      & \texttt{GCGCCACCGCGTATATTAGG} & \texttt{ATAATACGCGGTGAGTATAA} \\
      seq43840      & seq25057      & \texttt{GCGCCACCGCGTATATTAGG} & \texttt{TTCTAATATACGACCTGACA} \\
      seq43968      & seq59355      & \texttt{AAATAGCCTTTACTATGTCC} & \texttt{CTCGCAGTAAAGGCACCACC} \\
      \texttt{...}  & \texttt{...}  & \texttt{...}                  & \texttt{...}                  \\ \bottomrule
    \end{tabular}
    \label{tab:mmseqsres}
  \end{table}

  \noindent

  \section{Experimental platform}
  \label{sec-app:exp_platform}
  The various experiments in this study are performed on a number of different platforms to accommodate their unique requirements. Hyperparameter optimisation, general non-GPU (Graphics Processing Unit) intensive tasks and training of the simpler neural models (feed-forward neural networks) were performed on a portable computer equipped with a Core i9-8950HK processor, 32GB of DDR4 RAM, a PCI Express 3.0 solid-state drive (SSD) and an NVIDIA RTX 2070 external GPU connected through Thunderbolt 3, with 8GB VRAM.
  \\\\
  RoBERTa pre-training was performed on the Google Cloud Platform with 8 TPUv3 (third generation) cores and fine-tuned on a system equipped with 8 vCPUs, 61GB RAM, an SSD and an NVIDIA Tesla V100 with 16GB VRAM. Some compute-intensive tasks such as NUPACK or MMseqs2 were either run on our portable platform or an Azure virtual machine (VM) with 64 vCPUs and a premium SSD, as indicated in the main text. Due to the different configurations, we do not report \textit{training} times.
  \\\\
  Experiments involving the measurement of time were performed on three platforms. NUPACK code, which runs only on the Central Processing Unit (CPU) was timed on an Azure VM equipped with the hardware described above. The rest of the timing experiments were performed, by design, on consumer hardware. The two chosen consumer platforms are (1) a computer equipped with an AMD Ryzen 5950X CPU, 32GB of DDR4 RAM, a PCI Express 4.0 SSD and an NVIDIA RTX 3090 GPU with 24GB of VRAM and (2) Google Colaboratory with 8 TPUv2 cores (second generation).

  \section{Software}
  \label{apx:software}
  All code is written in the Python programming language and has been confirmed as working under revisions 3.8.5 and 3.8.6. The required machine learning libraries include scikit-learn 0.24, PyTorch 1.7.1 and PyTorch Lightning 1.1.4, which were the latest versions available as of January 2021. The main development operating system on the portable platform and the RTX 3090 GPU platform is Windows 10 Insider Preview Build 21286 (available on the Dev Channel), while the other platforms use various recent distributions of Linux.
  \\\\
  NUPACK, parasail and MMseqs2 computations were performed on the same portable computing platform as above, using their respective most recent versions as of May 2019. This translates to NUPACK version 3.2.2 and parasail versions 2.4.1 and 2.4.2. Our original CNN implementation (not presented in this paper) employed the most up-to-date TensorFlow 1 and Keras versions as of May 2019.

  \section{Measuring inference time}
  \label{sec-app:inf_time}
  The timing code for GPUs is based on the following snippet:

  \begin{minted}{python}
start = torch.cuda.Event(enable_timing=True)
end = torch.cuda.Event(enable_timing=True)

with torch.no_grad():
    start.record()
    trainer.test(model, test_dataloader)
    end.record()
    torch.cuda.synchronize()
    elapsed_time = start.elapsed_time(end)
\end{minted}

  \noindent The code can be trivially extended to perform multiple repetitions of the same code block in order to report the average execution time and standard deviation.
  \\\\
  \noindent Omitting the non-timing code, the timing code structure for TPUs is as follows:

  \begin{minted}{python}
class Module(pl.LightningModule):
    def __init__(self, ...):
        super(Module, self).__init__()
        self.forward_flag = 0

    def forward(self, x):
        if not self.forward_flag:
            self.forward_flag = 1
            self.start = time.time()

        x = ...

    def on_test_epoch_end(self):
        end = time.time()
        elapsed_time = end - self.start

\end{minted}

  \noindent This introduced probe effect will alter the execution time, but we assume the impact is negligible. Importantly, the above code is run on all of the 8 available TPU cores, meaning that each core reports its own time. As the cores operate concurrently, we take the mean of the 8 elapsed times and assign the result to the corresponding trial.
  \\\\
  It is also important to note that raw GPU performance roughly doubles with each generation, a trend also observed by Tensor Processing Units. Thus, the deep learning models we have developed will continue to improve in terms of inference time without any additional effort. A further optimisation would be the use of 16-bit precision. For a minor accuracy penalty, 16-bit implementations can both reduce video memory usage and improve execution times by several factors in the range of $\times1-5$.
  \\\\
  Furthermore, at the time of writing, the latest NVIDIA GPU architecture, Ampere, is still not fully exploited under CUDA Toolkit 11.0, the official version supported by PyTorch 1.7.1. For this reason, we compiled from source a nightly version of PyTorch 1.8.0 with CUDA 11.1 and this setup was used for all GPU timing experiments. However, it is likely that further optimisations will be possible on the Ampere architecture.

  \section{Detailed performance metrics on other temperatures}
  \label{apx-sec-other-temp}
  \cref{tab:other_temps_extended} presents the MSE, MCC and AUROC for the five evaluated temperatures and \cref{tab:prec_rec_f1_other_temps} lists per-class precision, recall and $\text{F}_{1}$ scores.

  \begin{table}[]
    \centering
    \caption{Summary of the evaluation metrics for the deep learning models at the selected temperatures.}
    \begin{tabular}{ccS[table-format=3.3]S[table-format=3.3]S[table-format=3.3]S[table-format=3.3]S[table-format=3.3]} \toprule
      \textbf{Model}                              & \textbf{Metric} & \textbf{37.0C} & \textbf{42.0C} & \textbf{47.0C} & \textbf{52.0C} & \textbf{62.0C} \\ \midrule
      \multirow{3}{*}{RNN}                        & MSE             & 808.126        & 655.260        & 433.125        & 190.772        & 333.637        \\
                                                  & MCC             & 0.862          & 0.872          & 0.888          & 0.916          & 0.833          \\
                                                  & AUROC           & 0.946          & 0.949          & 0.954          & 0.963          & 0.911          \\ \addlinespace
      \multirow{3}{*}{CNN}                        & MSE             & 719.616        & 576.134        & 374.768        & 170.865        & 398.881        \\
                                                  & MCC             & 0.884          & 0.893          & 0.908          & 0.929          & 0.811          \\
                                                  & AUROC           & 0.955          & 0.958          & 0.962          & 0.967          & 0.899          \\ \addlinespace
      \multirow{3}{*}{$\text{CNN}_{\text{Lite}}$} & MSE             & 712.598        & 574.295        & 381.676        & 188.943        & 418.637        \\
                                                  & MCC             & 0.893          & 0.901          & 0.914          & 0.929          & 0.801          \\
                                                  & AUROC           & 0.958          & 0.961          & 0.964          & 0.966          & 0.892          \\ \addlinespace
      \multirow{3}{*}{RoBERTa}                    & MSE             & 546.058        & 452.876        & 336.600        & 249.726        & 674.329        \\
                                                  & MCC             & 0.925          & 0.930          & 0.930          & 0.914          & 0.762          \\
                                                  & AUROC           & 0.971          & 0.971          & 0.968          & 0.954          & 0.870          \\ \bottomrule
    \end{tabular}
    \label{tab:other_temps_extended}
  \end{table}

  \begin{table}[t!]
    \centering
    \caption{Precision, recall and $\text{F}_{1}$ scores for the \textbf{Low} and \textbf{High} classes resulting from the deep learning models at the selected temperatures.}
    \begin{tabular}{ccccccccccc} \toprule
      \multicolumn{11}{c}{\textbf{CNNLite}}                                                                                                                                                                                                                           \\ \midrule
      \textbf{Metric}                & \multicolumn{2}{c}{\textbf{37.0C}} & \multicolumn{2}{c}{\textbf{42.0C}} & \multicolumn{2}{c}{\textbf{47.0C}} & \multicolumn{2}{c}{\textbf{52.0C}} & \multicolumn{2}{c}{\textbf{62.0C}}                                         \\ \cmidrule(lr){1-1} \cmidrule(lr){2-3} \cmidrule(lr){4-5} \cmidrule(lr){6-7} \cmidrule(lr){8-9} \cmidrule(lr){10-11}
      \textbf{Precision}             & 0.872                              & 0.998                              & 0.885                              & 0.996                              & 0.908                              & 0.992 & 0.947 & 0.979 & 0.999 & 0.817 \\
      \textbf{Recall}                & 0.996                              & 0.921                              & 0.993                              & 0.929                              & 0.987                              & 0.942 & 0.967 & 0.965 & 0.785 & 0.999 \\
      \textbf{$\text{F}_{\text{1}}$} & 0.930                              & 0.958                              & 0.936                              & 0.961                              & 0.946                              & 0.966 & 0.957 & 0.972 & 0.879 & 0.899 \\
                                     &                                    &                                    &                                    &                                    &                                    &       &       &       &       &       \\
      \multicolumn{11}{c}{\textbf{CNN}}                                                                                                                                                                                                                               \\ \midrule
      \textbf{Metric}                & \multicolumn{2}{c}{\textbf{37.0C}} & \multicolumn{2}{c}{\textbf{42.0C}} & \multicolumn{2}{c}{\textbf{47.0C}} & \multicolumn{2}{c}{\textbf{52.0C}} & \multicolumn{2}{c}{\textbf{62.0C}}                                         \\ \cmidrule(lr){1-1} \cmidrule(lr){2-3} \cmidrule(lr){4-5} \cmidrule(lr){6-7} \cmidrule(lr){8-9} \cmidrule(lr){10-11}
      \textbf{Precision}             & 0.860                              & 0.999                              & 0.873                              & 0.997                              & 0.897                              & 0.995 & 0.940 & 0.984 & 0.999 & 0.826 \\
      \textbf{Recall}                & 0.998                              & 0.912                              & 0.996                              & 0.920                              & 0.991                              & 0.934 & 0.975 & 0.960 & 0.798 & 0.999 \\
      \textbf{$\text{F}_{\text{1}}$} & 0.924                              & 0.953                              & 0.931                              & 0.957                              & 0.942                              & 0.963 & 0.957 & 0.971 & 0.887 & 0.905 \\
                                     &                                    &                                    &                                    &                                    &                                    &       &       &       &       &       \\
      \multicolumn{11}{c}{\textbf{RNN}}                                                                                                                                                                                                                               \\ \midrule
      \textbf{Metric}                & \multicolumn{2}{c}{\textbf{37.0C}} & \multicolumn{2}{c}{\textbf{42.0C}} & \multicolumn{2}{c}{\textbf{47.0C}} & \multicolumn{2}{c}{\textbf{52.0C}} & \multicolumn{2}{c}{\textbf{62.0C}}                                         \\ \cmidrule(lr){1-1} \cmidrule(lr){2-3} \cmidrule(lr){4-5} \cmidrule(lr){6-7} \cmidrule(lr){8-9} \cmidrule(lr){10-11}
      \textbf{Precision}             & 0.835                              & 0.999                              & 0.848                              & 0.998                              & 0.872                              & 0.996 & 0.918 & 0.988 & 0.999 & 0.844 \\
      \textbf{Recall}                & 0.998                              & 0.893                              & 0.997                              & 0.901                              & 0.993                              & 0.915 & 0.982 & 0.944 & 0.822 & 1.000 \\
      \textbf{$\text{F}_{\text{1}}$} & 0.909                              & 0.943                              & 0.917                              & 0.947                              & 0.929                              & 0.954 & 0.949 & 0.965 & 0.902 & 0.915 \\
                                     &                                    &                                    &                                    &                                    &                                    &       &       &       &       &       \\
      \multicolumn{11}{c}{\textbf{RoBERTa}}                                                                                                                                                                                                                           \\ \midrule
      \textbf{Metric}                & \multicolumn{2}{c}{\textbf{37.0C}} & \multicolumn{2}{c}{\textbf{42.0C}} & \multicolumn{2}{c}{\textbf{47.0C}} & \multicolumn{2}{c}{\textbf{52.0C}} & \multicolumn{2}{c}{\textbf{62.0C}}                                         \\ \cmidrule(lr){1-1} \cmidrule(lr){2-3} \cmidrule(lr){4-5} \cmidrule(lr){6-7} \cmidrule(lr){8-9} \cmidrule(lr){10-11}
      \textbf{Precision}             & 0.915                              & 0.995                              & 0.926                              & 0.992                              & 0.942                              & 0.983 & 0.962 & 0.957 & 0.996 & 0.788 \\
      \textbf{Recall}                & 0.991                              & 0.950                              & 0.986                              & 0.956                              & 0.971                              & 0.965 & 0.932 & 0.976 & 0.743 & 0.997 \\
      \textbf{$\text{F}_{\text{1}}$} & 0.951                              & 0.972                              & 0.955                              & 0.974                              & 0.956                              & 0.974 & 0.947 & 0.967 & 0.851 & 0.881 \\ \bottomrule
    \end{tabular}
    \label{tab:prec_rec_f1_other_temps}
  \end{table}

  \clearpage
  \section{Ablation study}
  \label{apx:abl-study}
  To study the effectiveness of the different extracted features, we perform an ablation study examining the classification performance of various configurations of interest. The results are presented in \cref{tab:abl-lda}, \cref{tab:abl-qda}, \cref{tab:abl-rf}, \cref{tab:abl-nn}. There are less entries for NN due to extended training time.

  \renewcommand{\arraystretch}{1.4}
  \begin{table}[H]
    \centering
    \caption{Ablation study summary for LDA.}
    \begin{tabular}{ccccccc} \toprule
      \textbf{Aln.} & \textbf{GC} & \textbf{S.C.} & \textbf{P.C.} & \textbf{S. MFE} & \textbf{MCC}   & \textbf{AUROC} \\ \midrule
      \checkmark    &             &               &               &                 & 0.840          & 0.906          \\
      \checkmark    & \checkmark  &               &               &                 & 0.846          & 0.909          \\
      \checkmark    & \checkmark  & \checkmark    &               &                 & 0.846          & 0.909          \\
      \checkmark    & \checkmark  &               & \checkmark    &                 & 0.846          & 0.909          \\
      \checkmark    & \checkmark  &               &               & \checkmark      & 0.850          & 0.912          \\
      \checkmark    & \checkmark  &               & \checkmark    & \checkmark      & 0.850          & 0.912          \\
      \checkmark    & \checkmark  & \checkmark    &               & \checkmark      & 0.850          & 0.912          \\
      \checkmark    & \checkmark  & \checkmark    & \checkmark    & \checkmark      & \textbf{0.851} & \textbf{0.912} \\
                    &             & \checkmark    &               &                 & 0.000          & 0.500          \\
                    &             &               & \checkmark    &                 & 0.000          & 0.500          \\
                    &             &               &               & \checkmark      & 0.000          & 0.500          \\
                    & \checkmark  & \checkmark    & \checkmark    & \checkmark      & 0.033          & 0.502          \\ \bottomrule
    \end{tabular}
    \label{tab:abl-lda}
  \end{table}

  \begin{table}[H]
    \centering
    \caption{Ablation study summary for QDA.}
    \begin{tabular}{ccccccc} \toprule
      \textbf{Aln.} & \textbf{GC} & \textbf{S.C.} & \textbf{P.C.} & \textbf{S. MFE} & \textbf{MCC}   & \textbf{AUROC} \\ \midrule
      \checkmark    &             &               &               &                 & 0.845          & 0.910          \\
      \checkmark    & \checkmark  &               &               &                 & 0.853          & 0.916          \\
      \checkmark    & \checkmark  & \checkmark    &               &                 & 0.852          & 0.918          \\
      \checkmark    & \checkmark  &               & \checkmark    &                 & 0.852          & 0.918          \\
      \checkmark    & \checkmark  &               &               & \checkmark      & 0.858          & 0.920          \\
      \checkmark    & \checkmark  &               & \checkmark    & \checkmark      & 0.859          & 0.922          \\
      \checkmark    & \checkmark  & \checkmark    &               & \checkmark      & 0.859          & 0.922          \\
      \checkmark    & \checkmark  & \checkmark    & \checkmark    & \checkmark      & \textbf{0.859} & \textbf{0.922} \\
                    &             & \checkmark    &               &                 & 0.023          & 0.504          \\
                    &             &               & \checkmark    &                 & 0.023          & 0.504          \\
                    &             &               &               & \checkmark      & 0.003          & 0.501          \\
                    & \checkmark  & \checkmark    & \checkmark    & \checkmark      & 0.096          & 0.548          \\ \bottomrule
    \end{tabular}
    \label{tab:abl-qda}
  \end{table}

  \noindent The table headers refer to the concepts introduced in \textit{Designing a diverse hybridisation dataset} (Aln., Alignment, GC, GC content, S.C., Single Concentration, P.C. Pair Concentration, S. MFE, Single MFE, MCC, Matthews Correlation Coefficient, AUROC, Area Under the Receiver Operating Characteristics).

  \begin{table}[H]
    \centering
    \caption{Ablation study summary for RF.}
    \begin{tabular}{ccccccc} \toprule
      \textbf{Aln.} & \textbf{GC} & \textbf{S.C.} & \textbf{P.C.} & \textbf{S. MFE} & \textbf{MCC}   & \textbf{AUROC} \\ \midrule
      \checkmark    &             &               &               &                 & 0.848          & 0.913          \\
      \checkmark    & \checkmark  &               &               &                 & 0.884          & 0.937          \\
      \checkmark    & \checkmark  & \checkmark    &               &                 & 0.875          & 0.934          \\
      \checkmark    & \checkmark  &               & \checkmark    &                 & 0.875          & 0.933          \\
      \checkmark    & \checkmark  &               &               & \checkmark      & \textbf{0.884} & \textbf{0.938} \\
      \checkmark    & \checkmark  &               & \checkmark    & \checkmark      & 0.879          & 0.935          \\
      \checkmark    & \checkmark  & \checkmark    &               & \checkmark      & 0.880          & 0.936          \\
      \checkmark    & \checkmark  & \checkmark    & \checkmark    & \checkmark      & 0.879          & 0.935          \\
                    &             & \checkmark    &               &                 & 0.141          & 0.547          \\
                    &             &               & \checkmark    &                 & 0.151          & 0.553          \\
                    &             &               &               & \checkmark      & 0.190          & 0.565          \\
                    & \checkmark  & \checkmark    & \checkmark    & \checkmark      & 0.559          & 0.780          \\ \bottomrule
    \end{tabular}
    \label{tab:abl-rf}
  \end{table}

  \begin{table}[H]
    \centering
    \caption{Ablation study summary for NN.}
    \begin{tabular}{ccccccc} \toprule
      \textbf{Aln.} & \textbf{GC} & \textbf{S.C.} & \textbf{P.C.} & \textbf{S. MFE} & \textbf{MCC}   & \textbf{AUROC} \\ \midrule
      \checkmark    & \checkmark  &               &               &                 & 0.867          & 0.931          \\
      \checkmark    & \checkmark  &               &               & \checkmark      & 0.872          & 0.932          \\
      \checkmark    & \checkmark  & \checkmark    & \checkmark    & \checkmark      & \textbf{0.873} & \textbf{0.930} \\ \bottomrule
    \end{tabular}
    \label{tab:abl-nn}
  \end{table}

  \noindent It is clear that alignment scores alone are a good predictor of hybridisation yield. A modest, but visible increase in classification performance is further provided by the quick-to-compute GC content. Depending on the machine learning model, the benefit of adding thermodynamic information is more or less subtle. In \cref{tab:abl-lda}, \cref{tab:abl-qda}, \cref{tab:abl-rf}, \cref{tab:abl-nn} the best scores are highlighted in bold: first by MCC; in case of equality also by AUROC; in case of equality of AUROC the most complete model (most features) is highlighted. For LDA, QDA and NN the best performing models use all features. For RF, the Aln-GC model has nearly identical performance to the Aln-GC-MFE model.
  \\\\
  However, the differences are statistically significant: for LDA, between the complete model and the Aln-GC variant ($P=.034$, permutation test with 5000 iterations, thus significant at the 95\% confidence interval); for QDA, between the complete model and the Aln-GC variant ($P<.001$, permutation test with 5000 iterations, thus significant at the 99\% confidence interval); for RF, between the Aln-GC-MFE model and the Aln-GC variant ($P<.001$, permutation test with 5000 iterations, thus significant at the 99\% confidence interval) and for NN between the complete model and the Aln-GC variant ($P<.001$, permutation test with 5000 iterations, thus significant at the 99\% confidence interval). The null hypothesis that the two groups come from the same distribution is rejected in all cases at the mentioned confidence interval. This two-sided permutation test is provided by the \texttt{permutation\_test} function of the mlxtend \cite{raschkas_2018_mlxtend} Python library. The statistical difference in otherwise close scores can be explained by trade-offs made by the model: some models might favour reducing false positives over false negatives and vice-versa.

  \section{Inference times for different dataset sizes}
  \label{apx:inf-diff-data-size}

\begin{figure}[h]
    \centering
    \includegraphics[width=1.0\textwidth]{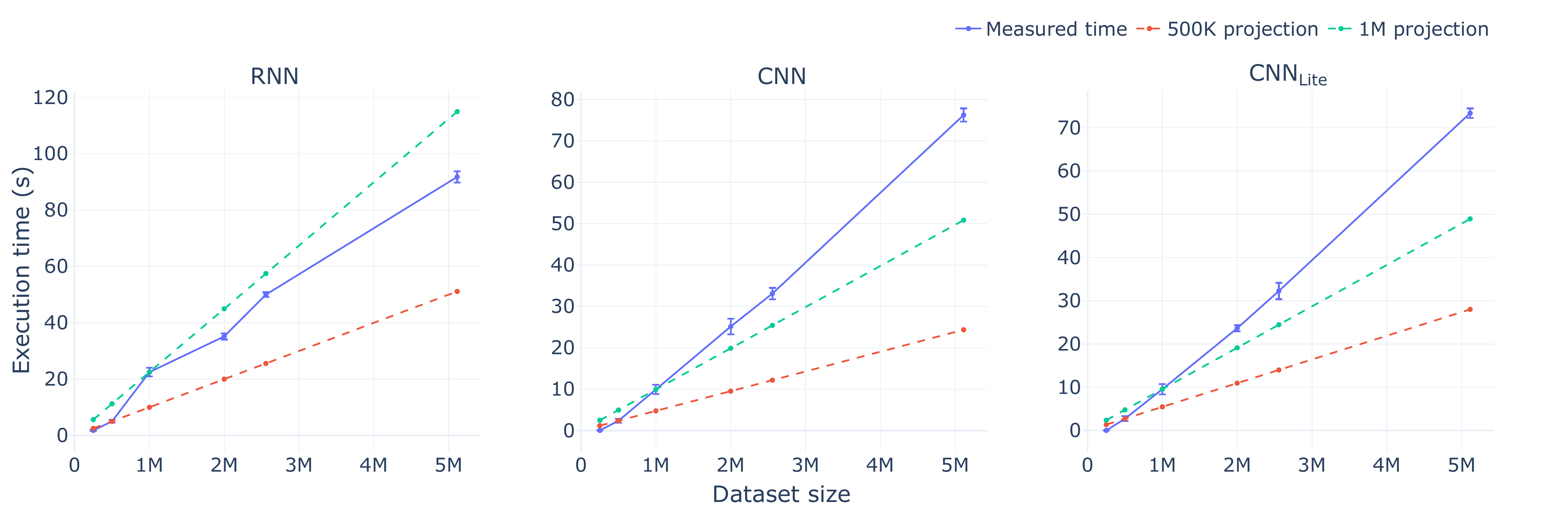}
    \caption{Scatter plots of the inference times for the three most efficient models: RNN, CNN and $\text{CNN}_{\text{Lite}}$ as the dataset size increases, together with projections based on the times for the 500K and 1M datasets. The batch size is set to \num[group-separator={,},group-minimum-digits={3}]{8192}. In the graphs, M is used for million(s).}
    \label{fig:apx-deep-learning-time}

  \end{figure}

  The inference time on second generation TPUs was measured using the methodology described in \Cref{sec-app:inf_time}, for subsets of size: \num[group-separator={,},group-minimum-digits={3}]{250000}, \num[group-separator={,},group-minimum-digits={3}]{500000}, \num[group-separator={,},group-minimum-digits={3}]{1000000}, \num[group-separator={,},group-minimum-digits={3}]{2000000} and \num[group-separator={,},group-minimum-digits={3}]{2556976} (full dataset). These datasets can be entirely loaded in memory, hence a PyTorch \texttt{TensorDataset} (without shuffling) was used to hold the data, initially loaded with NumPy \cite{harris2020array}. We further measured inference time on a dataset of size \num[group-separator={,},group-minimum-digits={3}]{5113952} (double the full dataset length), by concatenating two \texttt{TensorDataset} objects holding the 2.5 million data points using PyTorch's \texttt{ConcatDataset}. The batch size was set to \num[group-separator={,},group-minimum-digits={3}]{8192} as it enables better scalability on the larger datasets.
  \\\\
  \Cref{fig:apx-deep-learning-time} illustrates the measured inference time as well as the extrapolated values based on two reference points: \num[group-separator={,},group-minimum-digits={3}]{500000} and \num[group-separator={,},group-minimum-digits={3}]{1000000}, called "500K projection" and respectively "1M projection", for the three most efficient algorithms: RNN, CNN and $\text{CNN}_{\text{Lite}}$.
  \\\\
  All evaluated models see a spike in inference time for the 1 million dataset, however this trend improves for the RNN, whose times for the $>1$ million datasets stay below the 1M projected line (only $\times 1.84$ increase from 2.5M to 5M). The CNN and $\text{CNN}_{\text{Lite}}$ follow the 1M projected line closely until around 2.5M in dataset size. Afterwards, the difference compared to the 1M projected line increases; however the actual measured time increases by a factor of about $\times 2.27$ for $\text{CNN}_{\text{Lite}}$ and $\times 2.30$ for the CNN when transitioning from the 2.5M dataset to the 5M dataset. Thus, both the CNN and $\text{CNN}_{\text{Lite}}$ models are close the the ideal $\times 2$ increase in inference time as the dataset size doubles; in addition, the times are better than the RNN, which scales better than linearly relative to its own performance.

  \section{Similarity between yields at different temperatures}
  \label{apx:sec-yield-temps}

\begin{figure}
    \captionsetup[subfigure]{labelformat=empty}
    \begin{subfigure}[t]{.49\textwidth}
      \centering
      \includegraphics[width=1.0\textwidth]{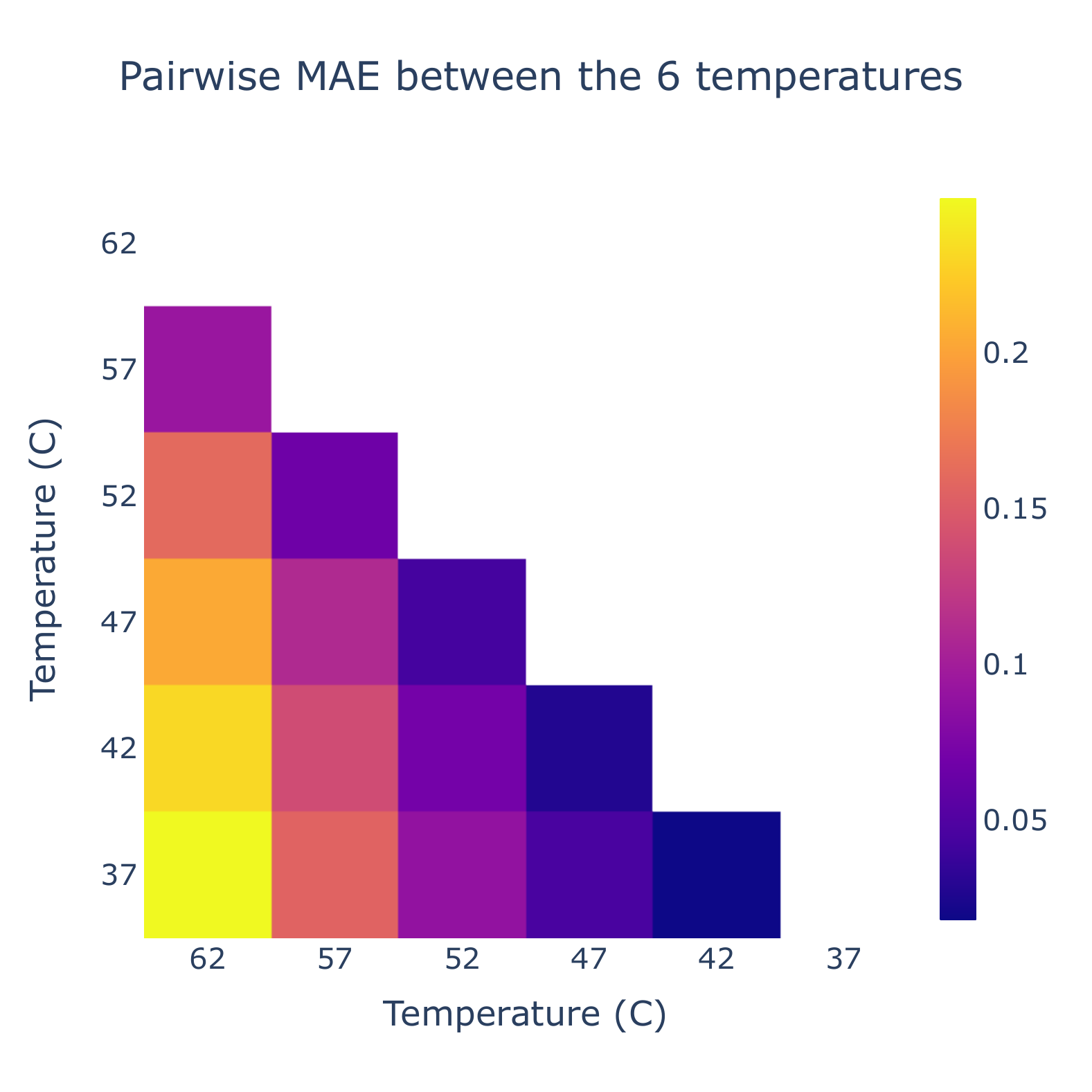}
    \end{subfigure}
    \begin{subfigure}[t]{.49\textwidth}
      \centering
      \includegraphics[width=1.0\textwidth]{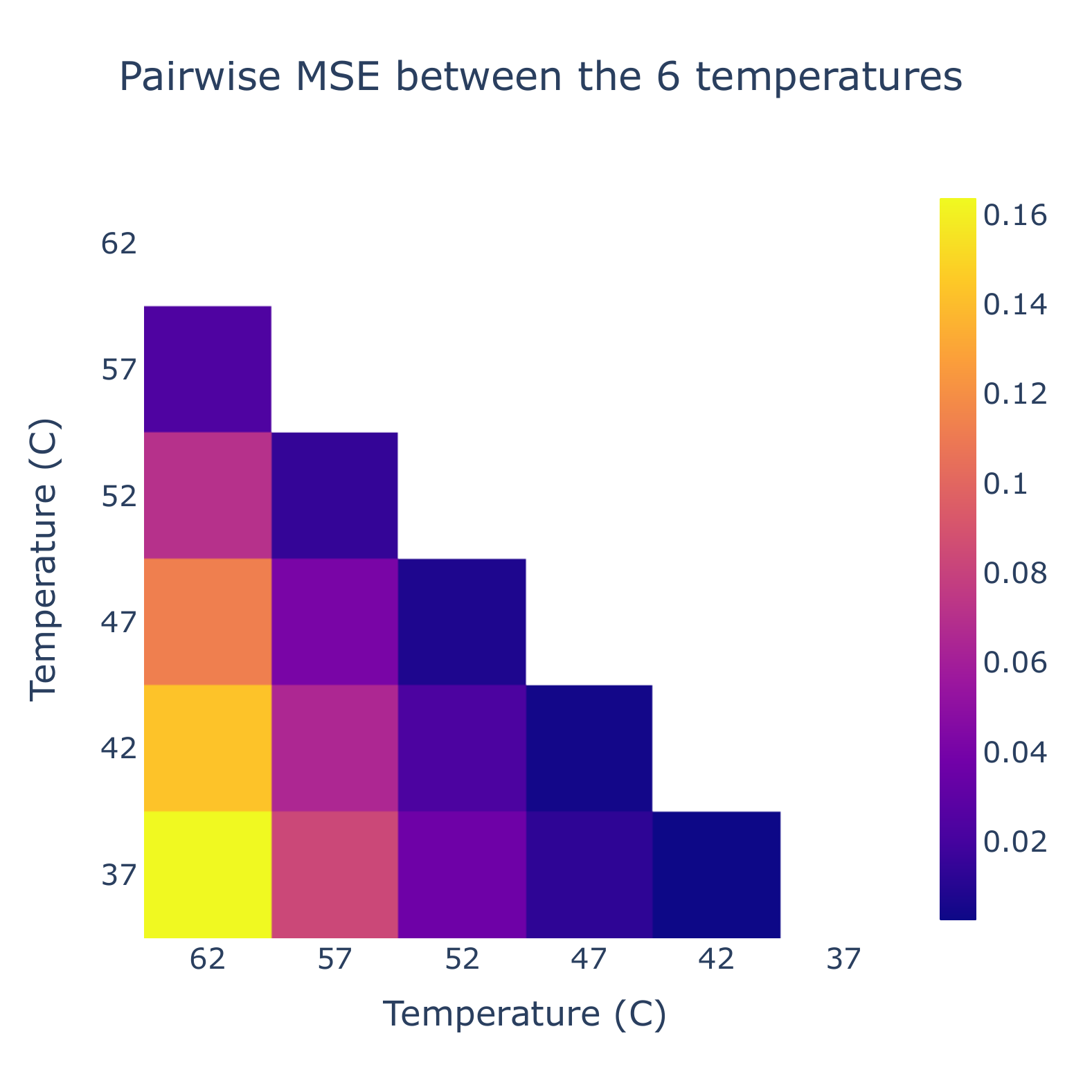}
    \end{subfigure}
    \caption{The ground truth yields at the six different temperatures are compared in a pairwise manner to assess their relatedness, using the MAE and MSE as the scoring functions.}
    \label{fig:yield-pairwise-matrix}
  \end{figure}

  An intuition for the similarity between the ground truth yields as computed by NUPACK at different temperatures can be attained by pairwise scoring the values using a criterion such as mean absolute error (MAE) or mean squared error (MSE). For this comparison, all yields are in the range $[0, 1]$. \Cref{fig:yield-pairwise-matrix} captures these metrics and indicates that the yields at 62\textdegree C are the most different, even when compared to their closest neighbour (57\textdegree C). On the other hand, yields in the range 37\textdegree C to 52 \textdegree C are relatively close to each other (in all cases MAE lower than 0.1, MSE lower than 0.04).
  \\\\
  The choice of 57\textdegree C for training and evaluation is motivated by being close to the standard melting and/or annealing temperatures of PCR primers (more details in the main text) and not being too far away from the behaviours at 62\textdegree C and $<57\degree$C as illustrated in \Cref{fig:yield-pairwise-matrix}.

\end{appendices}
%TC:endignore

\end{document}